\documentclass{article}

\usepackage{arxiv}

\usepackage{times}  
\usepackage{helvet}  
\usepackage{courier}  
\usepackage[hyphens]{url}  
\usepackage{graphicx} 
\usepackage[square, numbers]{natbib}  
\usepackage{caption} 
\usepackage{algorithm}
\usepackage{algorithmic}

\usepackage{newfloat}
\usepackage{listings}
\DeclareCaptionStyle{ruled}{labelfont=normalfont,labelsep=colon,strut=off} 
\floatstyle{ruled}
\newfloat{listing}{tb}{lst}{}
\floatname{listing}{Listing}

\title{Physics-Informed Time-Integrated DeepONet: Temporal Tangent Space Operator Learning for High-Accuracy Inference}

\author{
 Luis Mandl\\
 Institute of Structural Mechanics and Dynamics in\\
 Aerospace Engineering, University of Stuttgart;\\
  Experimental Hepatobiliary Surgery Group,\\
  Department of Hepatobiliary Surgery\\
  and Visceral Transplantation,\\
  University of Leipzig Medical Center;\\
  Department of Civil and Systems Engineering,\\
  Johns Hopkins University\\
  \texttt{luis.mandl@isd.uni-stuttgart.de} \\
   \And
 Dibyajyoti Nayak\\
  Department of Civil and Systems Engineering,\\
  Johns Hopkins University\\
  \texttt{dnayak2@jh.edu} \\
  \And
 Tim Ricken\\
  Institute of Structural Mechanics and Dynamics in\\
  Aerospace Engineering, University of Stuttgart\\
  \texttt{tim.ricken@isd.uni-stuttgart.de} \\
  \And
 Somdatta Goswami\\
  Department of Civil and Systems Engineering,\\
  Johns Hopkins University\\
  \texttt{somdatta@jhu.edu} \\
}

\usepackage{bibentry}

\usepackage{comment}
\usepackage{multirow,booktabs}
\usepackage{fixme}
\usepackage{amsmath}
\usepackage{subcaption}
\usepackage{amsfonts} 
\usepackage{hyperref}
\usepackage{booktabs}

\newcommand{\revresponse}[1]{#1}

\DeclareMathSymbol{\shortminus}{\mathbin}{AMSa}{"39}
\newcommand\eminus{\mathrm{e}\text{-}}
\begin{document}

\maketitle
\begin{abstract}
Accurately modeling and inferring solutions to time-dependent partial differential equations (PDEs) over extended temporal horizons remains a core challenge in scientific machine learning. Traditional full rollout (FR) methods, predicting entire trajectories in a single pass, often fail to capture the causal dependencies inherent to dynamical systems and exhibit poor generalization outside the training time horizon. In contrast, autoregressive (AR) approaches, which evolve the system step by step, are prone to error accumulation, as predictions at each time step depend on potentially erroneous prior outputs. These shortcomings limit the long-term accuracy and reliability of both strategies. To overcome these issues, we introduce Physics-Informed Time-Integrated Deep Operator Network (PITI-DeepONet), an operator learning framework designed for stable and accurate long-term time evolution, well beyond the training time horizon. PITI-DeepONet employs a dual-output DeepONet architecture trained via either fully physics-informed or hybrid physics- and data-driven objectives. The training enforces consistency between the learned temporal derivative and its counterpart obtained via automatic differentiation. Rather than directly forecasting future states, the network learns the time-derivative operator from the current state, which is then integrated using classical time-stepping schemes - such as explicit Euler, fourth-order Runge-Kutta, second-order Adams-Bashforth-Moulton\revresponse{, or implicit Euler} - to advance the solution sequentially in time. Additionally, the framework \revresponse{supports} residual monitoring during inference to estimate prediction quality and \revresponse{flags when the learned temporal tangent becomes unreliable, e.g., outside the training domain}. Applied to benchmark problems, PITI-DeepONet demonstrates enhanced accuracy and stability over extended inference time horizons when compared to traditional methods. Mean relative $\mathcal{L}_2$ errors reduced by 84\% (versus FR) and 79\% (versus AR) for the one-dimensional heat equation; by 87\% (versus FR) and 98\% (versus AR) for the one-dimensional Burgers equation; by 42\% (versus FR) and 89\% (versus AR) for the two-dimensional Allen-Cahn equation\revresponse{; and by 58\% (vs. FR) and 61\% (vs. AR) for the one-dimensional Kuramoto-Sivashinsky equation}. By moving beyond classic FR and AR schemes, PITI-DeepONet paves the way for more reliable, long-term integration of complex, time-dependent PDEs.
\end{abstract}


\section{Introduction}
Neural operators (NOs) have emerged as powerful surrogate models in recent years, achieving remarkable success in both forward and inverse problems
~\cite{lu2021learning,li2021fourier, Kobayashi2024, kag2024learning}. These include applications in additive manufacturing~\cite{Liu2024, Kushwaha2024}, nuclear energy systems~\cite{Kobayashi2024b}, and biomedical topics~\cite{goswami2022neural} to name a few. Physics-informed approaches~\cite{Karniadakis2021, Wang2021, Goswami2023} have also gained prominence, enabling the training process to dispense entirely, or in part, with labeled data pairs by relying instead on physical laws and residual minimization, akin to solving classical initial-boundary value problems of partial differential equations~(PDEs). However, the way operator learning is employed differs fundamentally from traditional numerical methods for solving differential equations. Common approaches include full rollout (FR) models, which train on an entire spatiotemporal domain and then perform inference on that same domain~\cite{lin2023learning, wang2023long}.

While the FR approach avoids treating time evolution explicitly, it often struggles with extrapolation beyond the training domain and fails to leverage the Markovian nature of many dynamical systems, where subsequent states inherently depend on previous ones. Alternatively, discrete autoregressive rollout (AR) methods have been introduced but face the limitations of training domain generalization and an additional susceptibility to error accumulation in time-dependent systems~\cite{mccabe2023stabilityautoregressiveneuraloperators}. As a result, both approaches are ill-suited for making reliable long-term predictions akin to those obtained via classical simulations, limiting their utility as surrogate models.

Several recent efforts have sought to overcome these limitations by introducing architectural modifications in the NO to provide long-time prediction capabilities. For instance, in ~\cite{diab2025temporal}, an additional network was introduced to better capture temporal dependencies, while NO outputs were processed further by augmenting recurrent architectures to improve sequence modeling~\cite{michalowska2024neural}. Other approaches introduce memory mechanisms~\cite{buitrago2024benefits, he2024sequential} to mitigate the non-Markovian behavior inherent in rollout-based methods. Nonetheless, these methods predominantly adopt generic time-series solutions without exploiting the rich dynamical information present in time-dependent systems. 

The time-integration embedded operator learning framework introduced~\cite{2024Nayak-TI-DeepONet} addresses these limitations by leveraging the system’s intrinsic dynamical structure, enabling near real-time inference, and shifting part of the computational load to an offline phase. This framework learns the discrete tangent from data and reuses it during inference through a numerical time integrator. Building on this foundation, we propose a significant advancement: a continuous temporal tangent space operator, which generalizes the notion of learned time evolution beyond discrete updates, transitioning the training process from a purely data-driven approach to a combination of physics-informed and data-driven methodologies. Additionally, we incorporate a residual prediction mechanism based on the current system state, which serves as an effective error predictor. This online residual tracking provides a means to assess prediction quality with minimal computational overhead, offering new opportunities for robust and efficient operator learning.

\section{Methodology}
\subsection{Time-Integrator embedded Deep Operator Network}
The Time-Integrator embedded Deep Operator Network (TI-DeepONet) framework~\cite{2024Nayak-TI-DeepONet} transforms the task of temporal prediction in dynamical systems into a derivative learning problem. Rather than predicting future states in an autoregressive manner, the model learns a discrete approximation of the system's temporal tangent space. Concretely, the current system state $\mathbf{u}^n$ is passed through a DeepONet~\cite{lu2021learning}, which outputs the estimated temporal derivative $\hat{\mathbf{u}}_t^n$. This derivative is then integrated numerically via schemes like Euler or Runge-Kutta to yield the predicted next state $\hat{\mathbf{u}}^{n+1}$. The model is trained using a supervised loss computed between $\hat{\mathbf{u}}^{n+1}$ and the true state $\mathbf{u}^{n+1}$ and learns an approximate tangent between the data pairs via the underlying scheme as a discrete operator. This setup tightly couples operator learning with time integration and provides a modular structure, where physical priors and numerical schemes can be systematically incorporated.

\subsection{Physics-Informed Time-Integrated Deep Operator Network}
\revresponse{In contrast, the proposed physics-informed time-integrated deep operator network (PITI-DeepONet) employs a physics-informed ansatz to learn a continuous temporal tangent operator that maps the current state directly to its time derivative. While TI-DeepONet learns a discrete, fixed-$\Delta t$ tangent map that is tied to the training time step and uses learnable coefficients to compensate when integrating with a chosen numerical scheme, PITI-DeepONet decouples operator learning from time integration: once the continuous tangent is learned, inference can use any standard time integrator with arbitrary step sizes \(\Delta t\) without requiring learnable integration weights.} In the following, we provide a comprehensive overview of how the physics-informed continuous tangent space operator is derived, along with a discussion on the training process, data sampling, and inference setup. \revresponse{We want to emphasize that PITI is an ansatz-level construction and therefore operator-agnostic, i.e., it can be combined with any neural operator backbone as long as a physics-informed instantiation of that operator is feasible.}
\begin{figure}[htb!]
    \centering
    \includegraphics[width = 0.8\linewidth]{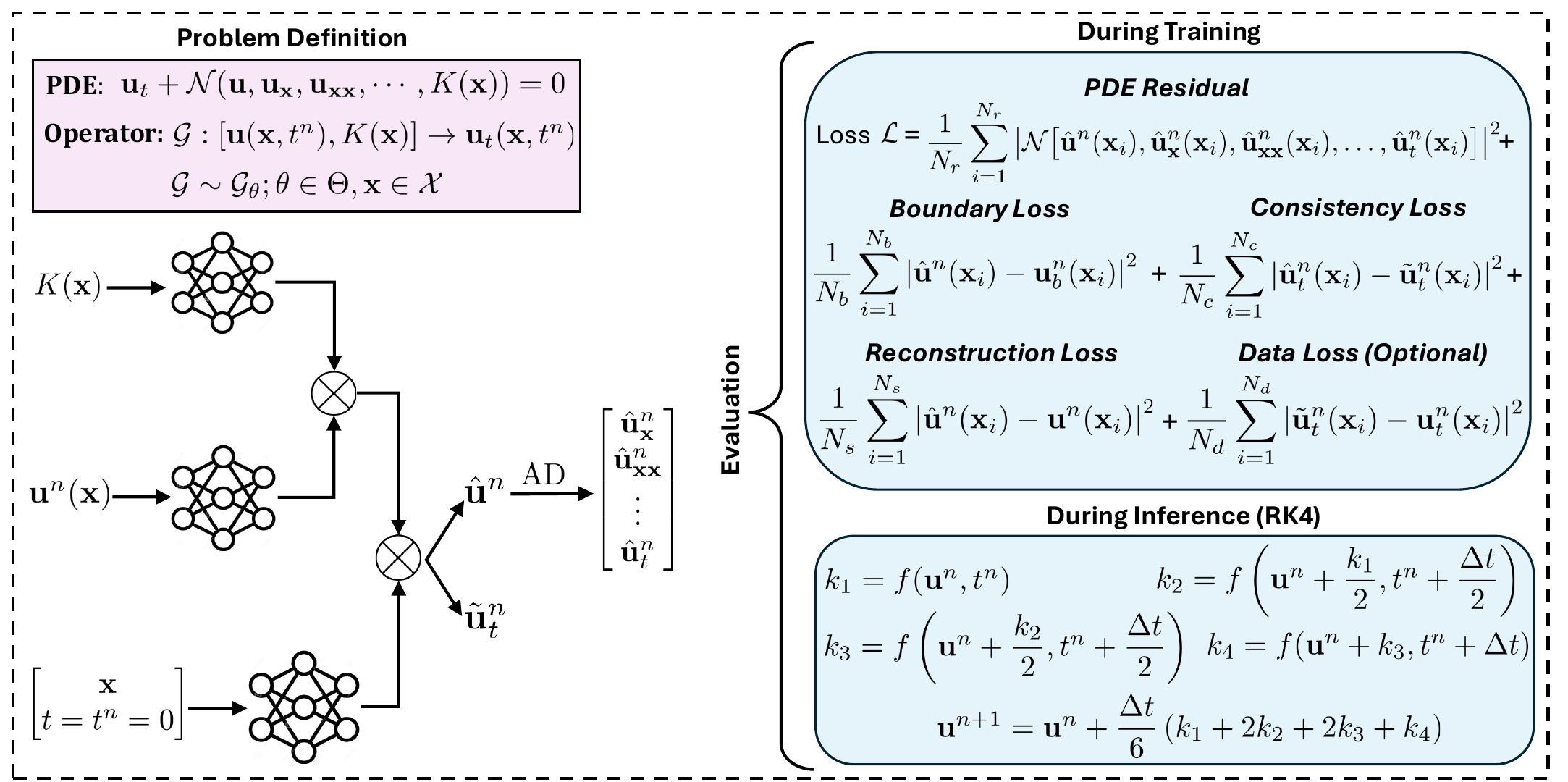}
    \caption{A schematic of the proposed physics-informed time-integrated deep operator network (PITI-DeepONet) architecture, which learns a continuous temporal tangent space operator that enables efficient and accurate time-stepping during inference.}
    \label{fig:PI-TI-DON-schema}
\end{figure}

\subsubsection{Network and Training Setup}\label{sec:training_setup}
To learn the temporal tangent ($\mathbf{u}_t^n$), we use a physics-informed DeepONet that performs the mapping $\mathcal{G}:\mathbf{u}^n\rightarrow{\hat{\mathbf{u}}}^n, \widetilde{{\mathbf{u}}}^n_t$, where $\mathbf{u}^n$ is the input field, $\hat{\mathbf{u}}^n$ is the reconstructed input field, and ${\widetilde{\mathbf{u}}}^n_t$ is the predicted temporal tangent. For physics-informed training, we use automatic differentiation (AD) on $\hat{\mathbf{u}}^n$ to obtain the derivatives necessary to compute the loss terms: (1) PDE residual, $\mathcal{L}_{\rm{PDE}}$, (2) initial condition (IC) loss, $\mathcal{L}_{\rm{IC}}$, and (3) boundary condition (BC) loss, $\mathcal{L}_{\rm{BC}}$. This part is identical to physics-informed training of any operator~\cite{Wang2021}. In addition, we introduce two new losses as follows. First, the discrepancy between $\mathbf{u}^n$ and $\hat{\mathbf{u}}^n$ is quantified as reconstruction loss $\mathcal{L}_{\rm{R}}$. Second, the consistency of the temporal derivative with the field is ensured using a consistency loss, $\mathcal{L}_{\mathrm{C}}$, between the network-predicted derivative, $\widetilde{\mathbf{u}}_t^n$, and the temporal derivative obtained via automatic differentiation (AD) on $\hat{\mathbf{u}}^n$, i.e., $\hat{\mathbf{u}}_t^n$. To accommodate the temporal dimension in this formulation, phantom values (e.g., zeros) are included as placeholders in the time input to the trunk network (to indicate the current time), alongside other coordinate inputs. Nonetheless, this still effectively reduces the dimensionality of the underlying problem to some extent as only a single value for the time domain has to be sampled. A schematic of the architecture described above is presented in Fig.~\ref{fig:PI-TI-DON-schema}. Moreover, as shown in the schematic, data-driven losses can be additionally imposed on the network outputs to achieve a hybrid setup. \revresponse{Importantly, PITI does not inherently require time-derivative ground truth: the data-driven term $\mathcal{L}_{\rm{u_t}}$ is optional and is used to stabilize and speed up training in this implementation, while ground-truth $\mathbf{u}_t$ is additionally required to quantify temporal-tangent errors on held-out test data. In our benchmarks, $\mathbf{u}_t$ targets are obtained by evaluating the known PDE right-hand side at stored snapshots (no additional PDE solves; for explicit schemes, this quantity is already computed during time stepping). Therefore, the additional computational cost is minimal compared to generating the reference solution trajectories, which AR and FR training also require.} In total, this leads to 6 loss terms that are aggregated as the total loss. Note that all losses are computed as mean squared errors (MSE) and then combined with weighting factors $\mathbf{\lambda}=\{\lambda_{\rm{PDE}},\lambda_{\rm{R}}, \lambda_{\rm{BC}}, \lambda_{\rm{C}}, \lambda_{\rm{u}}, \lambda_{\rm{u_t}}\}$. The total loss reads $\mathcal{L} = \lambda_{\rm{PDE}} \mathcal{L}_{\rm{PDE}} + \lambda_{\rm{R}} \mathcal{L}_{\rm{R}} + \lambda_{\rm{BC}} \mathcal{L}_{\rm{BC}}+\lambda_{\rm{C}} \mathcal{L}_{\rm{C}} +\lambda_{\rm{u}} \mathcal{L}_{\rm{u}} + \lambda_{\rm{u_t}} \mathcal{L}_{\rm{u_t}}$, with the latter two being the data-driven losses, which are optional. Furthermore, the reconstruction loss and the data-driven loss for $\mathbf{u}^n$ have the same formal structure, but are assigned different weights within the loss function to balance the influence of the data-driven and physics-informed terms. This allows for a purely physics-informed or a hybrid setup. All models are optimized using ADAM~\cite{Adam2014} with an exponential decay schedule for the learning rate. Note that there are other ways of formulating the PITI-DeepONet, but each have their disadvantages compared to the proposed method: i) one could discard the output, ${\widetilde{\mathbf{u}}}^n_t$, and employ only AD to obtain the time derivative in inference. However, this would significantly increase the computational demand and reduce inference speed; ii) one could skip the consistency loss and directly incorporate the output, ${\widetilde{\mathbf{u}}}^n_t$, in the PDE residual. \revresponse{This becomes the special formulation for explicit time-evolution PDEs of the form $u_t=\mathcal{N}(\cdot)$, where the time derivative appears in isolation. In this case, we can form the PDE residual by inserting the predicted tangent $\widetilde{\mathbf{u}}_t^n$ directly into the PDE loss, which avoids any automatic differentiation with respect to $t$ and reduces the need for the consistency loss. Consequently, the trunk does not require an explicit time input, and no dummy $0$ placeholders are needed. Nevertheless}, as will be shown later in this work, the resulting network output is comparable or even inferior to our setup, so the more general formulation is preserved.

\subsubsection{Sampling}\label{sec:sampling} Key to learning a temporal tangent space operator is to span the domain of solution states during inference with the limited samples used during training. Unlike other methods, which typically involve time evolution as seen in the traditional use of operators, this approach focuses solely on the time derivative of the current state. Consequently, a more meticulous sampling strategy is required. One potential remedy is to increase both the number and diversity of randomly sampled ICs to effectively span the relevant solution manifold. However, achieving sufficient coverage of the solution space through random sampling alone can be daunting—especially in the absence of domain knowledge to guide and reduce the training data requirements. In this work, we adopt a different strategy that leverages available domain knowledge in the form of solution samples $\mathbf{u}^n$ and corresponding time derivatives ${\mathbf{u}}^n_t$ at fixed time instances. These samples may originate from experiments, analytical solutions, numerical simulations, or even pre-trained surrogate models operating over longer time horizons. Once obtained, they are aggregated and used as inputs to the branch network, treated equivalently to IC samples. The trunk network, in turn, receives these inputs with phantom zeroes in the time dimension, omitting explicit time encoding, alongside spatial coordinates and any additional parameters. \revresponse{As PITI is queried at the current state during rollout, ensuring reliable performance necessitates that the training branch inputs encompass the state distribution anticipated at inference. Therefore, the sampling strategy cannot be completely agnostic: it needs to be constructed using information specific to the problem (such as expected IC and parameter ranges) to guarantee that the learned temporal tangent is accurate on the states observed during inference, instead of targeting a single universal operator applicable to any arbitrary unseen regimes.}

\subsubsection{Inference Setup}
\label{sec:inference-setup}
Having learned a continuous temporal tangent space operator, we can now exploit this for time-stepping-based inference. For each available state as branch input, the network predicts a replica (or reconstruction) of the input state as well as the corresponding temporal tangent (or time-derivative) of that state. To advance the solution in time, we utilize well-established, stable, and accurate time-stepping schemes commonly used for solving ordinary differential equations (ODEs). There are no constraints imposed on parameters such as model complexity, choice of numerical method, number of iterations, or time step size. \revresponse{Since PITI learns a continuous temporal tangent, the subsequent time integration is decoupled from the training setup and can be performed with any numerical scheme during inference, with accuracy governed by the integrator’s standard stability and order properties.} To illustrate the flexibility of our approach, \revresponse{the following} representative numerical time integration schemes are considered.

\noindent First, the explicit Euler method is used, which is a first-order single-step scheme. Given $\mathbf{u}^n$ at time step $n$ and its time derivative $\widetilde{\mathbf{u}}_t^n$ as obtained from the learned operator, the solution is advanced according to
\begin{equation}
\mathbf{u}^{n+1} = \mathbf{u}^{n} + \Delta t \,\widetilde{\mathbf{u}}_t^n = \mathbf{u}^{n} + \Delta t \,\mathcal{G}_{\boldsymbol{\theta}}(\mathbf{u}^n).
\end{equation}
Second, the fourth-order Runge--Kutta method (RK4) is employed, representing a classical multi-stage scheme with improved accuracy through intermediate evaluations: 
\begin{align}
k_1 &= \mathcal{G}_{\boldsymbol{\theta}}(\mathbf{u}^n), \\
k_2 &= \mathcal{G}_{\boldsymbol{\theta}}\left(\mathbf{u}^n + \tfrac{\Delta t}{2}k_1\right), \\
k_3 &= \mathcal{G}_{\boldsymbol{\theta}}\left(\mathbf{u}^n + \tfrac{\Delta t}{2}k_2\right), \\
k_4 &= \mathcal{G}_{\boldsymbol{\theta}}\left(\mathbf{u}^n + \Delta t\,k_3\right), \\
\mathbf{u}^{n+1} &= \mathbf{u}^n + \frac{\Delta t}{6}\left(k_1 + 2k_2 + 2k_3 + k_4\right),
\end{align}
where $k_1$, $k_2$, $k_3$, and $k_4$ are the intermediate RK4 slopes.

\noindent Third, the second-order Adams-Bashforth-Moulton method (ABM2) is included, functioning as a two-step predictor-corrector scheme that utilizes information from multiple time levels:
\begin{align}
\text{Predictor:} \quad \widetilde{\mathbf{u}}^{n+1} &= \mathbf{u}^n + \Delta t\left(\tfrac{3}{2}\,\mathcal{G}_{\boldsymbol{\theta}}(\mathbf{u}^n) - \tfrac{1}{2}\,\mathcal{G}_{\boldsymbol{\theta}}(\mathbf{u}^{n-1})\right), \\
\text{Corrector:} \quad \mathbf{u}^{n+1} &= \mathbf{u}^n + \frac{\Delta t}{2}\left(\mathcal{G}_{\boldsymbol{\theta}}(\mathbf{u}^n) + \mathcal{G}_{\boldsymbol{\theta}}\left(\widetilde{\mathbf{u}}^{n+1}\right)\right)
\end{align}

\noindent \revresponse{Lastly, the backward Euler method (BE) is included as an A-stable implicit method. The initial guess is formed using explicit Euler, and the next guesses are computed by solving a non-linear fixed-point relation with damped Picard iterations:
\begin{align}
\text{Initialization:}\quad \mathbf{u}^{n+1,(0)} 
&= \mathbf{u}^{n} + \Delta t \,\mathcal{G}_{\boldsymbol{\theta}}\!\left(\mathbf{u}^n\right),\\
\text{Picard map:}\quad \widetilde{\mathbf{u}}^{n+1,(k+1)} 
&= \mathbf{u}^n + \Delta t\,\mathcal{G}_{\boldsymbol{\theta}}\!\left(\mathbf{u}^{n+1,(k)}\right),\\
\text{Relaxation:}\quad \mathbf{u}^{n+1,(k+1)} 
&= (1-\omega)\,\mathbf{u}^{n+1,(k)} + \omega\,\widetilde{\mathbf{u}}^{n+1,(k+1)}, 
\qquad k=0,\dots,K-1.
\end{align}
where $\omega\in[0,1]$ is a tunable damping parameter and $k$ the fixed number of iterations. Within this work, we use $K=16$ iterations with $\omega=0.4$.}

It should be noted that in the above numerical schemes employed during inference, $\mathcal{G}_{\boldsymbol{\theta}}$ represents the learned operator, parametrized by $\boldsymbol{\theta}$, mapping the input state to its time derivative ($\widetilde{\mathbf{u}}_t^n$). Only the time derivative output is used to perform numerical time integration. These integration schemes span a wide range of numerical properties in terms of stability, accuracy, and computational cost, making them suitable for evaluating the behavior of the proposed framework under varying conditions. We further distinguish between the training time step $\mathrm{dt}$ (used implicitly through sampled data) and the inference time step $\Delta t$ (used explicitly during numerical integration), offering additional flexibility in evaluating generalization and stability behavior.

Another notable property of this method is the ability to monitor the residual, $\mathcal{R}(\mathbf{u}^n, \hat{\mathbf{u}}^n) = (\mathbf{u}^n - \hat{\mathbf{u}}^n)^2$, computed using the input state $\mathbf{u}^n$ and the network prediction $\hat{\mathbf{u}}^n$. In theory, a well-learned state should correspond to a small residual, indicating consistency between the model's prediction and the governing dynamics. Conversely, a large residual may suggest that the current input lies outside the distribution of the training data, potentially leading to inaccurate predictions by the learned operator. This behavior effectively provides an intrinsic estimate of prediction quality, enabling a form of self-assessment within the inference process. \revresponse{As a large residual indicates that the learned time derivative may be unreliable, continuing the rollout would propagate incorrect derivative inputs, and a practical deployment could switch to an alternative derivative source, e.g., a different surrogate or a conventional solver. In this work, we use residual monitoring solely for reporting and do not adapt the inference procedure based on $\mathcal{R}$.}

\subsection{Reference Methods: Full Rollout and Autoregressive Rollout}
\label{sec:reference-methods}
To demonstrate the advantages of the proposed approach, we present a comprehensive comparison against state-of-the-art methods, including FR and AR. For FR, the operator is trained on the temporal domain, $t \in [0, 0.5]$ for all examples, learning to predict the entire spatiotemporal output field over this interval in a single shot (without respecting causality). In AR, the training process uses three temporal slices per trajectory, specifically at $t \in \lbrace 0, 0.25, 0.5 \rbrace$, to learn discrete mappings from each slice to the subsequent step. Within hybrid physics-informed and data-driven formulations, data losses for FR are applied continuously over the temporal domain, while in AR, they are localized to the selected time slices, with both being supplemented by physics-informed losses. \revresponse{The physics-informed collocation is paradigm-dependent: FR enforces the residual over the full training horizon $t\in[0,T]$, whereas AR enforces it only over a single-step window starting from the current time $t^n$ (equivalently acting on a local interval of length $\Delta t$, i.e., $[0,\Delta t]$, per transition), and PITI enforces it at the current-time mapping only (equivalently at a single time point, i.e., $[0]$). In hybrid settings, the supervised targets are likewise method-specific: FR uses trajectory snapshots, AR uses state pairs $(\mathbf{u}^n,\mathbf{u}^{n+1})$, and PITI uses time-derivative targets $\mathbf{u}_t^n$, but in all cases they are derived from the same reference trajectories for identical initial conditions, ensuring a fair comparison.} This design ensures a consistent and equitable comparison between FR and AR by providing both methods with access to identical training information. Note that the losses here are PDE, IC, BC, and a data loss, where IC refers to the reconstruction loss but is named IC to better reflect its purpose in numerical settings.
\revresponse{Beyond FR and AR, we incorporate two additional baselines from the literature for 1D Burgers’ equation. First, we consider physics-constrained deep autoregressive networks (PCAR; \cite{Geneva2020}), which advance the solution by repeatedly applying a state-to-state convolutional predictor that maps the current (and optionally several past) spatial fields to the next field with a fixed time step. Architecturally, this predictor is a dense encoder–decoder convolutional neural network consisting of encoding convolutions, a dense block, and a decoding block that upsamples to the original grid. Training is purely physics-constrained by minimizing a discretized time-integrator residual (e.g., Crank–Nicolson / Euler variants), with spatial derivatives approximated via finite-difference-like stencil convolutions, thereby eliminating the need for labeled trajectory pairs or automatic differentiation. To obtain a purely physics-informed PCAR baseline under the same training data and hyperparameter budget as our AR and PITI variants, we retrained PCAR on our dataset and adapted it to our setup by using a single input slice instead of multiple previous slices.}
\revresponse{Second, we include the physics-informed neural ODE (PINODE; \cite{Sholokhov2023}), a reduced-order modeling framework that jointly learns an autoencoder (encoder–decoder pair) and a neural model of time evolution in a low-dimensional latent space. Specifically, the autoencoder compresses the physical state into latent variables, a neural network governs their temporal evolution, and an ODE solver integrates this evolution forward in time before decoding back to the original state. Its physics-informed training enforces consistency with the governing dynamics via collocation: the learned latent evolution is regularized so that, when passed through the encoder/decoder, it matches the known physical evolution while maintaining accurate reconstruction, which necessitates a differentiable encoder and numerically stable automatic differentiation. In contrast to discrete autoregressive rollouts, PINODE operates intrinsically in continuous time, so its long-horizon behavior is primarily determined by the latent dimensionality, the expressiveness of the latent dynamics network, and the numerical stability of the ODE integration, rather than by the number of rollout steps used in training. For PINODE, we report only the published results for the 1D Burgers’ equation (rather than re-implementing the method, as no code was provided alongside the work) and use these as an external reference for long-horizon rollout accuracy in the literature.}

\section{Numerical Results}

\begin{table*}[!ht]
    \centering
    \caption{Minimum, mean, and maximum relative $\mathcal{L}_2$ error along with mean and standard deviation for 5 timed runs each over all test examples for \revresponse{Heat, Burgers', Allen-Cahn, and Kuramoto-Sivashinsky~(KS) equation} in comparison of FR, AR, and time integration using explicit Euler, fourth-order Runge-Kutta~(RK4), second-order Adam-Bashforth-Moultons~(ABM2) \revresponse{and implicit Euler (IE)} as inference scheme with $\Delta t = 0.01$. Irrespective of the inference scheme employed, PITI‑DeepONet decreases the mean relative $\mathcal{L}_2$ errors by 84\% (vs. FR) and 79\% (vs. AR) for the one‑dimensional heat equation; by 87\% (vs. FR) and 98\% (vs. AR) for the one‑dimensional Burgers' equation; by 42\% (vs. FR) and 89\% (vs. AR) for the two‑dimensional Allen–Cahn equation\revresponse{; and by 58\% (vs. FR) and 61\% (vs. AR) for the one-dimensional Kuramoto-Sivashinsky equation}. The timing includes residual calculation for PITI cases.}
    \setlength{\tabcolsep}{3.9pt}
    \footnotesize{
    \resizebox{\columnwidth}{!}{\begin{tabular}{ll|cc|cc|cc|cc}
    \toprule
        && \multicolumn{2}{c|}{Heat 1D @ $t=5.0$} & \multicolumn{2}{c|}{Burgers 1D @ $t=1.0$} & \multicolumn{2}{c|}{Allen-Cahn 2D @ $t=1.0$} & \multicolumn{2}{c}{\revresponse{KS 1D @ $t=10.0$}} \\ \midrule
        && Min\,/\,Mean\,/\,Max & Inference in $s$  & Min\,/\,Mean\,/\,Max & Inference in $s$ & Min\,/\,Mean\,/\,Max & Inference in $s$ & \revresponse{Min\,/\,Mean\,/\,Max} & \revresponse{Inference in $s$} \\\cmidrule{3-10}
        \multicolumn{2}{c|}{FR} & 0.82\,/\,1.5\,/\,24 & $1.4\eminus 3\!\pm\!1.7\eminus 5$ & 0.048\,/\,0.13\,/\,0.56 & $5.8\eminus 4\!\pm\!3.1\eminus 5$ & 0.13\,/\,0.20\,/\,0.28 & $9.5\eminus 4\!\pm\!1.7\eminus 5$ & \revresponse{0.71\,/\,1.11\,/\,2.55} & \revresponse{$8.6\eminus4\!\pm\!2.6\eminus5$}\\
        \multicolumn{2}{c|}{AR} & 0.11\,/\,1.2\,/\,23 & $6.8\eminus 2\!\pm\!3.9\eminus 5$ & 0.15\,/\,0.94\,/\,11 & $1.6\eminus 3\!\pm\!7.6\eminus 5$ & 0.49\,/\,1.1\,/\,1.7 & $9.2\eminus 3\!\pm\!2.2\eminus 4$ & \revresponse{0.27\,/\,1.20\,/\,2.17} & \revresponse{$7.0\eminus3\pm3.2\eminus5$}\\\midrule
        \multirow{4}{*}{\rotatebox[origin=c]{90}{PITI}} & Euler & 0.029\,/\,0.24\,/\,3.4 & $7.5\eminus 2\!\pm\!1.7\eminus 5$ & 0.0084\,/\,0.018\,/\,0.26 & $1.4\eminus 2\!\pm\!9.0\eminus 5$ & 0.079\,/\,0.12\,/\,0.16 & $8.9\eminus 3\!\pm\!3.1\eminus 5$ & \revresponse{0.074\,/\,0.46\,/\,1.36} & \revresponse{$1.4\eminus2\pm2.6\eminus5$}\\
        & RK4 & 0.029\,/\,0.24\,/\,3.3 & $2.0\eminus 1\!\pm\!2.6\eminus 4$ & 0.0038\,/\,0.017\,/\,0.26 & $3.8\eminus 2\!\pm\!8.5\eminus 5$ & 0.079\,/\,0.12\,/\,0.16 & $2.6\eminus 2\!\pm\!3.7\eminus 5$ & \revresponse{-} & \revresponse{-}\\
        & ABM2 & 0.029\,/\,0.24\,/\,3.3 & $1.7\eminus 2\!\pm\!2.2\eminus 4$ & 0.0039\,/\,0.017\,/\,0.26 & $3.2\eminus 2\!\pm\!3.6\eminus 5$ & 0.079\,/\,0.12\,/\,0.16 & $2.4\eminus 3\!\pm\!6.1\eminus 5$ & \revresponse{-} & \revresponse{-}\\
         & IE & \revresponse{-} & \revresponse{-} & \revresponse{-} & \revresponse{-} & \revresponse{-} & \revresponse{-} & \revresponse{0.083\,/\,0.47\,/\,1.42} & \revresponse{$1.4\pm7.7\eminus2$}\\ \bottomrule
    \end{tabular}
    }}
    \setlength{\tabcolsep}{6pt}
    \label{tab:result_overview}
\end{table*}
\begin{figure}[htp]
  \centering
  \begin{subfigure}[b]{0.24\textwidth}
    \centering
    \includegraphics[width=\textwidth]{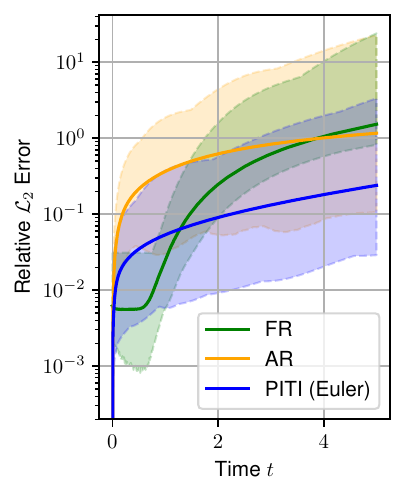}
    \caption{Heat 1D}
    \label{fig:heat_l2}
  \end{subfigure}
  \hfill
  \begin{subfigure}[b]{0.24\textwidth}
    \centering
    \includegraphics[width=\textwidth]{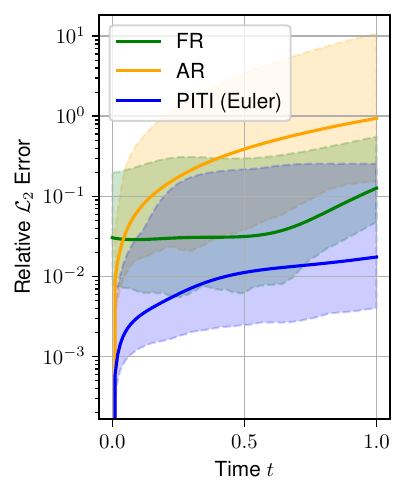}
    \caption{Burgers 1D}
    \label{fig:burgers_l2}
  \end{subfigure}
  \hfill
  \begin{subfigure}[b]{0.24\textwidth}
    \centering
    \includegraphics[width=\textwidth]{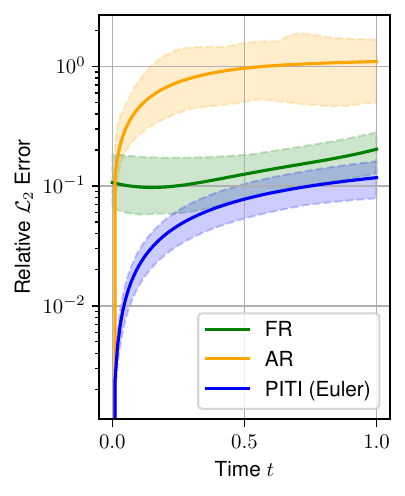}
    \caption{Allen-Cahn 2D}
    \label{fig:ac_l2}
  \end{subfigure}
  \hfill
  \begin{subfigure}[b]{0.24\textwidth}
    \centering
    \includegraphics[width=\textwidth]{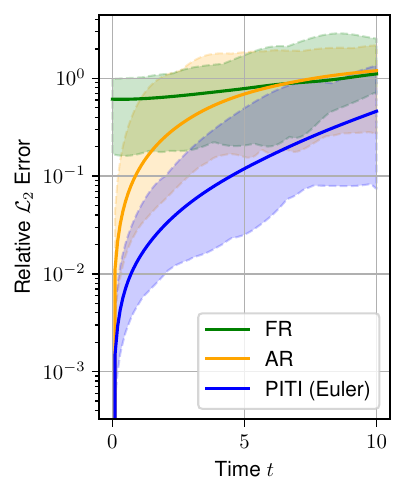}
    \caption{\revresponse{Kuramoto-Sivashinsky 1D}}
    \label{fig:ks_l2}
  \end{subfigure}

  \caption{Relative $\mathcal{L}_2$ error over time for the \revresponse{four} PDEs in comparison of FR, AR, and time integration using explicit Euler with $\Delta t = 0.01$. Depicted are the mean, as well as the minimum and the maximum values, for all test examples. Only inference with explicit Euler is depicted, but all methods show similar values in inference, also as shown in Tab. \ref{tab:result_overview}.}
  \label{fig:result_overview}
\end{figure}
The subsequent sections detail the setup and results for the examples of Heat and Burgers' equations in one dimension and Allen-Cahn equation in two dimensions \revresponse{and close with Kuramoto–Sivashinsky (KS) equation in 1D as an example for chaotic behaviour}. Fig. \ref{fig:result_overview} and Tab. \ref{tab:result_overview} present a short overview of the performance of our proposed method compared to FR and AR. The relative $\mathcal{L}_2$ error = $\|\hat{\mathbf{u}}^n - \mathbf{u}^n\|_2 / \|\mathbf{u}^n\|_2$ was employed on the predicted field as standard metric for comparison with the field, depicting relative deviation from the ground truth. Furthermore, all hyperparameters were optimized using \textit{optuna}~\cite{akiba2019optuna}. The metric for AR and FR models was to reduce the relative $\mathcal{L}_2$ error on the complete validation data for the field output, while PITI models were optimized on the relative $\mathcal{L}_2$ error of the time derivative within the complete validation data. Optimization was performed on a discrete set of pre-chosen options to ease interpretation. At least 20 runs with a minimum of 6h run time were performed per setup. This included depth and width of branch and trunk nets, hidden dimension, training epochs, learning rate settings, loss weights, as well as information on how to obtain multiple outputs by splitting either or both trunk and branch outputs where necessary. For networks with CNNs in the branch, the number of layers, kernels per layer, depth, and size of the dense layers, if pooling was applied, and if so, whether average or max pooling, as well as dense and CNN activation. Stride for convolutions and pooling, as well as kernel size for convolutions, were fixed to 2 in each dimension. Interestingly, all CNNs converged to the same structure, including activation functions, except for the width of the dense layer. The results are given in Tab. \ref{tab:hyperparameters} for everything except loss weights, which are presented in Tab. \ref{tab:loss_weights}. Random seed choice in physics-informed scientific machine learning can influence solution diversity and optimization outcomes, especially in problems with multiple valid solutions, since random initialization may lead to discovering different solutions~\cite{zou2025learningdiscoveringmultiplesolutions}. However, as the obtained results showed good agreement during training with the used input, a single random seed per numerical example was utilized. This is a standard, resource-efficient practice within the field, but we acknowledge that additional seeds might yield alternative results in non-convex or multi-solution settings. Training and inference were performed on an \textit{NVIDIA A40} GPU with 48 GB of memory, utilizing CUDA version 12.4 and NVIDIA driver version 550.54.15, with the GPU operating in a Linux environment running the 64-bit Debian-based GNU/Linux operating system under kernel version 6.1.0-34-cloud-amd64. The framework was built in \textit{Python} 3.11.2 based on \textit{JAX}~\cite{jax2018github} and \textit{Flax}~\cite{flax2020github} with integration of tools from the \textit{DeepMind Ecosystem}~\cite{deepmind2020jax}. The full code included utilized package versions will be available at \url{https://github.com/lmandl/PITI-DeepONet} upon publication of this work. The dataset for the underlying cases was generated from scratch for this paper. While some datasets exist and either the data or the code to generate the dataset are publicly available~(cf. \cite{Wang2021} for Burger's equation), so far no dataset includes high-accuracy time derivatives. Hence, the need for this data to track the accuracy of our method or additional input for hybrid data-driven and physics-informed approaches required the generation of new datasets that fit the purpose of solving PDEs.
\begin{table*}[tbph]
    \centering
    \caption{Hyperparameter values after optuna optimization for all setups. Branch inputs are 128 sensors for the heat equation \revresponse{and Kuramoto-Sivashinsky~(KS) equation}, 101 sensors for Burgers' equation, and $(16,16)$ for the Allen-Cahn equation. Trunk input dimensions are 2 for the heat\revresponse{,} Burgers'  \revresponse{and KS} equation~$(t,x)$ and 3 for the Allen-Cahn equation~$(t, x, y)$. For models with the special form for time-evolution PDEs, the trunk input is reduced by the time dimension. p denotes the latent dimension for the branch and trunk nets. All models are unstacked DeepONet models that are optimized with an Adam optimizer and exponential learning rate decay. \textit{Hy.} indicates a hybrid data-driven and physics-informed model, while \textit{spec.} indicates the special form. The architecture of the convolutional neural network~(CNN) used for all three Allen-Cahn~(AC) cases consists of an initial 2D convolution~(Conv) with a stride of 2 and 2 filters, followed by average pooling~(AvgPool) with a stride of 2, then another 2D convolution with a stride of 2 and 4 filters, and another average pooling with a stride of 2. The output is then flattened and passed through a dense layer, with the size specified in the column for the branch network. Activation function for all models, including the CNNs, is the hyperbolic tangent\revresponse{, unless stated otherwise}. Timing is averaged over the whole training, including 100 validation evaluations equally distributed throughout training.}
    \footnotesize{
    \resizebox{\columnwidth}{!}{\begin{tabular}{ll|llll|lll|ll|c}
    \toprule
         & & & & & & \multicolumn{3}{c|}{exp. lr decay} & \multicolumn{2}{c|}{split} \\
        \multicolumn{2}{c|}{Model} & branch & trunk & p & epochs & base & rate & steps & branch & trunk & IT/s\\
        \midrule
        \parbox[t]{2mm}{\multirow{4}{*}{\rotatebox[origin=c]{90}{Heat}}} 
          & PITI       & 8x[128]   & 10x[128]   & 256  & 300,000 & $1\eminus 4$  & 0.9  & 40,000  & False & True & 151.76 \\
          & FR       & 8x[64]    & 8x[32]     & 64   & 300,000 & $5\eminus 4$  & 0.85 & 50,000  & False & False & 150.95 \\
          & AR       & 4x[256]   & 10x[256]   & 64   & 300,000 & $5\eminus 5$  & 0.95 & 10,000  & False & False & 171.08 \\
          & PITI \emph{spec.} & 8x[128]   & 8x[256]    & 128  & 300,000 & $5\eminus 5$  & 0.8  & 100,000 & False & True  & 309.59 \\ 
        \midrule
        \parbox[t]{2mm}{\multirow{5}{*}{\rotatebox[origin=c]{90}{Burgers}}} 
          & PITI       & 10x[256]  & 8x[64]     & 64   & 500,000 & $1\eminus 4$  & 0.9  & 100,000 & False & True & 206.33 \\
          & PITI \emph{Hy.}   & 8x[128]   & 8x[128]    & 128  & 300,000 & $1\eminus 4$  & 0.95 & 5,000   & True  & True & 395.96 \\
          & AR \emph{Hy.}   & 6x[256]   & 10x[256]   & 256  & 200,000 & $5\eminus 5$  & 0.9  & 10,000  & False & False & 338.76 \\ 
          & FR \emph{Hy.}   & 8x[128]   & 6x[64]     & 256  & 100,000 & $1\eminus 3$  & 0.85 & 7,500 & False & False & 168.23 \\ 
          & PITI \emph{spec.} & 6x[64]    & 10x[32]    & 128  & 200,000 & $5\eminus 5$  & 0.8  & 20,000  & True  & False & 625.04 \\ 
        \midrule
        \parbox[t]{2mm}{\multirow{3}{*}{\rotatebox[origin=c]{90}{AC}}}     
          & PITI \emph{Hy.}   & CNN*, [256]& 6x[32]     & 128  & 50,000  & $1\eminus 3$  & 0.9  & 100,000 & False & True & 129.88 \\ 
          & AR \emph{Hy.}   & CNN*, [64] & 6x[128]    & 256  & 100,000 & $1\eminus 3$  & 0.95 & 7,500 & False & False & 192.64\\ 
          & FR \emph{Hy.}   & CNN*, [64] & 6x[128]    & 128  & 25,000  & $1\eminus 4$  & 0.95 & 40,000  & False & False & 22.14 \\ 
        \midrule
        \parbox[t]{2mm}{\multirow{3}{*}{\rotatebox[origin=c]{90}{\revresponse{KS}}}} 
          & \revresponse{PITI \emph{Hy.}}   & \revresponse{10x[64]}   & \revresponse{6x[256]\textsuperscript{\textdagger}}    & \revresponse{256}  & \revresponse{200,000} & \revresponse{$1\eminus 4$}  & \revresponse{0.85} & \revresponse{30,000}   & \revresponse{False}  & \revresponse{True} & \revresponse{259.77} \\
          & \revresponse{AR \emph{Hy.}}   & \revresponse{4x[64]}   & \revresponse{4x[64]}   & \revresponse{128}  & \revresponse{200,000} & \revresponse{$5\eminus 4$}  & \revresponse{0.8}  & \revresponse{20,000}  & \revresponse{False} & \revresponse{False} & \revresponse{564.62} \\ 
          & \revresponse{FR \emph{Hy.}}   & \revresponse{10x[64]}   & \revresponse{4x[256]}   & \revresponse{256}  & \revresponse{50,000} & \revresponse{$5\eminus 4$}  & \revresponse{0.9} & \revresponse{50,000} & \revresponse{False} & \revresponse{False} & \revresponse{23.83} \\ 
    \bottomrule
    \addlinespace[0.25em]
    \multicolumn{12}{l}{* CNN consists of Conv2D[2], AvgPool, Conv2D[4], AvgPool, where Conv2D[$j$] indicates $j$ filters in that layer}\\
    \multicolumn{12}{l}{\revresponse{\textsuperscript{\textdagger} Sine as trunk activation function}}
\end{tabular}}
    }
    \label{tab:hyperparameters}
\end{table*}
\begin{table}[tbph]
    \centering
    \caption{Loss weights after optuna optimization for all setups. Factors for reconstruction loss in time integration and initial condition loss in rollout settings are combined into one column, as they essentially have the same significance.}
    \vskip1.25mm
    \footnotesize{
    \begin{tabular}{ll|lllllll|c}
    \toprule
         &  & \multicolumn{6}{c}{Loss Weights}\\
        \multicolumn{2}{c|}{Model} & $\lambda_{\rm{PDE}}$ & $\lambda_{\rm{R}}$ & $\lambda_{\rm{BC}}$ & $\lambda_{\rm{u}}$ & $\lambda_{\rm{u_t}}$ & $\lambda_{\rm{C}}$ \\ \midrule
        \parbox[t]{2mm}{\multirow{4}{*}{\rotatebox[origin=c]{90}{Heat}}}
          & PITI       & 1   & 10   & 1   & -   & -   & 1  \\
          & FR       & 1   & 2.5  & 1   & -   & -   & -  \\
          & AR       & 10  & 100  & 1   & -   & -   & -   \\
          & PITI spec. & 10  & 20   & 1   & -   & -   & -   \\ \midrule
        \parbox[t]{2mm}{\multirow{5}{*}{\rotatebox[origin=c]{90}{Burgers}}}
          & PITI       & 2   & 10   & 1   & -   & -   & 4  \\
          & PITI Hy.   & 1   & 1    & 20  & 2   & 50  & 20  \\
          & AR Hy.   & 5   & 10   & 1   & 25  & -   & -  \\
          & FR Hy.   & 25  & 100  & 10  & 1   & -   & -   \\
          & PITI spec. & 2   & 20   & 1   & 10  & 20  & -  \\ \midrule
        \parbox[t]{2mm}{\multirow{3}{*}{\rotatebox[origin=c]{90}{AC}}}
          & PITI Hy.   & 2.5 & 2.5  & 1   & 5   & 25  & 10  \\
          & AR Hy.   & 10  & 1    & 2   & 10  & -   & -  \\
          & FR Hy.   & 10  & 50   & 1   & 10  & -   & -  \\ \midrule
        \parbox[t]{2mm}{\multirow{3}{*}{\rotatebox[origin=c]{90}{\revresponse{KS}}}}
          & \revresponse{PITI Hy.}  & \revresponse{1} & \revresponse{10}  & \revresponse{10}   & \revresponse{10}   & \revresponse{1000}  & \revresponse{10}  \\
          & \revresponse{AR Hy.}   & \revresponse{1}  & \revresponse{50}    & \revresponse{50}   & \revresponse{50}  & \revresponse{-}   & \revresponse{-}  \\
          & \revresponse{FR Hy.}   & \revresponse{50}  & \revresponse{1}   & \revresponse{2.5}   & \revresponse{50}  & \revresponse{-}   & \revresponse{-}  \\
    \bottomrule
\end{tabular}
    }
    \label{tab:loss_weights}
\end{table}
\subsection{One-dimensional Heat Equation}
As an introductory example, we present the 1D heat equation in the form:
\begin{equation}
\label{eq:heat}
    \frac{\partial T}{\partial t} = 0.01~\frac{\partial^2 T}{\partial x^2},\quad (t, x)\in[0,\infty)\times [0,1]
\end{equation} with Dirichlet BC $T(t,0)=T(t,1)=0$. IC, $T(0,x)=f(x)$, is sampled from a periodic Gaussian process with $f(0)=f(1)=0$, length scale = $0.1$, and variance $\sigma^2=10,000$. The spatial domain was discretized using 128 grid points. The training was performed with a purely physics-informed setup using 4,800 input profiles sampled from 1,600 ICs at the time steps $t = \lbrace0, 0.25, 0.5\rbrace$. Note our remarks in the section on sampling (Sec.~\ref{sec:sampling}), as this does not change the underlying problem but rather represents a method for sampling the required profiles to cover the domain. The discrete time step for AR is set to $\rm{dt}=0.01$, while FR training was performed on $t \in [0,0.5]$ with the same timestep. Additionally, 400 validation samples on $t\in[0,1]$ and 500 test samples on $t\in[0,5]$, both with time step $\rm{dt}=0.01$ were used.

The dataset is generated by solving Eq. \ref{eq:heat} with Dirichlet BCs on the mentioned spatiotemporal domain. A uniform grid with $N_x = 128$ points is used for space, and the solution is computed over $N_{t,\mathrm{fine}} = 10,000$ fine time steps with a step size $\Delta t_{\mathrm{fine}}$, which inherently satisfies the stability requirements of the numerical scheme. The IC $u_0(x)$ is sampled from a periodic Gaussian random field. The numerical solution is obtained using an explicit finite-difference scheme, with second-order accurate spatial derivatives in the interior and fourth-order one-sided differences at the boundaries. The time derivative $\partial_t u$ is computed at each step, and BCs are enforced by setting $u(t, 0) = u(t, 1) = 0$. The solution is then downsampled to a coarse temporal grid with $N_t = 100$ intervals using linear interpolation, yielding the coarse fields $u(t, x)$ and $\partial_t u(t, x)$. The dataset includes $N_{\mathrm{train}} = 2,000$ samples for training and $N_{\mathrm{test}} = 500$ for testing, with solutions recorded for $T = 1$ and $T_{\mathrm{test}} = 5$, respectively.

\begin{figure}[htb!]
    \centering
    \includegraphics[width=0.49\linewidth]{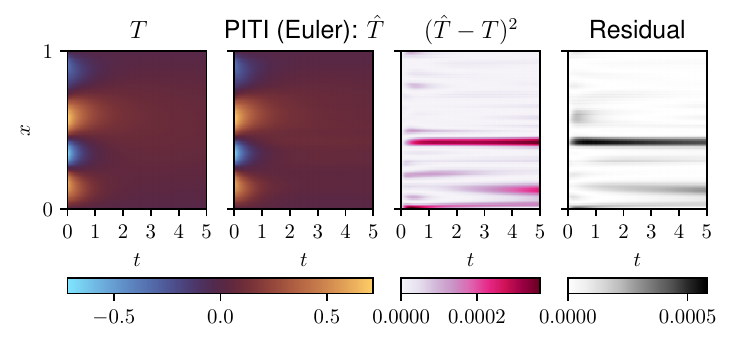}
    \caption{1D Heat equation (Mean rel. $\mathcal{L}_2$ error $7.60\eminus 2$): Comparison of reference data and PITI-DeepONet with explicit Euler alongside the squared difference of the computed field and the residuals during inference. Mean absolute error between squared difference and predicted residual for this example is $3.72\eminus 5$ and Pearson's correlation coefficient is $\rho=0.9985$.}
    \label{fig:heat_example}
\end{figure}

\begin{figure}[tbhp]
    \centering
    \includegraphics[width=\linewidth]{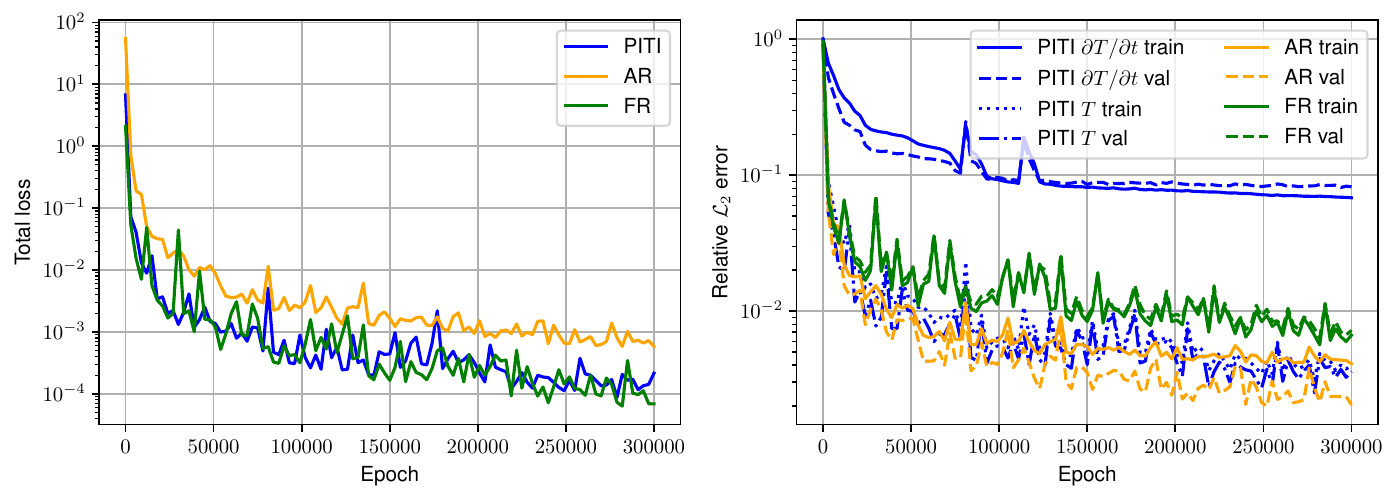}
    \caption{1D Heat equation: Total loss during training alongside mean relative $\mathcal{L}_2$ error on the train and validation set for full rollout (FR), autoregressive (AR), and time integration (PITI) model. The relative $\mathcal{L}_2$ error on both network outputs is provided for the PITI model. Hyperparameters were adapted per model as per table \ref{tab:hyperparameters}.}
    \label{fig:heat_loss}
\end{figure}

\begin{figure}[tbhp]
    \centering
    \includegraphics[width=0.49\linewidth]{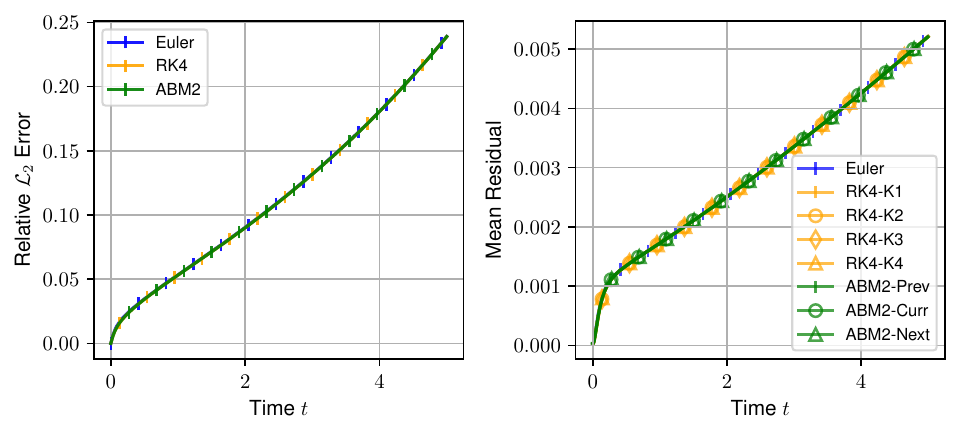}
    \caption{1D Heat equation: Comparison of different time integration (PITI) methods using explicit Euler, fourth-order Runge-Kutta (RK4) and second-order Adam-Bashforth-Moultons (ABM2) with mean relative $\mathcal{L}_2$ error on the left and predicted residuals on the right over all test examples. \revresponse{Note that in both subplots several curves overlap almost exactly and therefore appear as a single trace.}}
    \label{fig:heat_ti_inference}
\end{figure}

\begin{figure}[tbhp]
    \centering
    \includegraphics[width=0.49\linewidth]{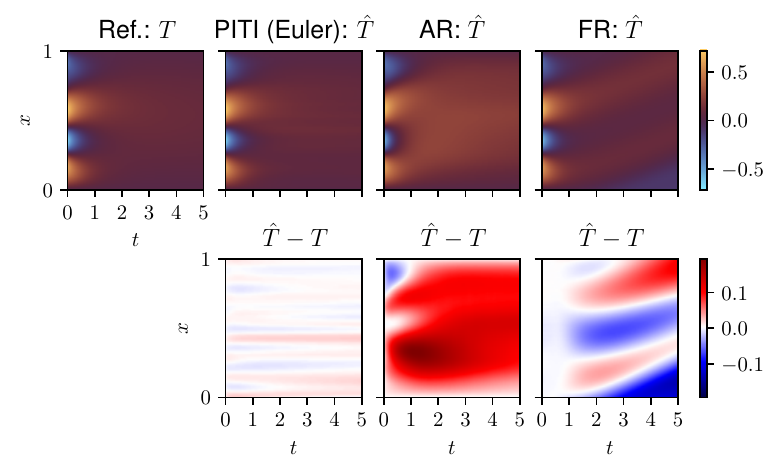}
    \caption{1D Heat equation: Comparison of predictions made using time integration (PITI) with explicit Euler, autoregressive, and full rollout alongside the differences for each method.}
    \label{fig:heat_field_plots}
\end{figure}

Final mean relative $\mathcal{L}_2$ errors observed were $6.81\eminus 2$ on the training set and $8.20\eminus 2$ on the validation set for the time derivative using TI, whereas for the field values, AR achieved $4.09\eminus 3$ and $2.04\eminus 3$, and FR achieved $6.63\eminus 3$ and $7.20\eminus 3$ on the training and test sets, respectively~(see Fig.~\ref{fig:heat_loss} for full trajectories). Although FR and AR achieve an order of magnitude better results on the training domain $t \in [0,1]$, if used in inference, PITI beats AR on $t \in [0,1]$ and both AR and FR on $t \in [0,5]$ as shown in Fig. \ref{fig:heat_l2} and Tab. \ref{tab:result_overview} with full field plots and differences for a single test example for PITI using explicit Euler, AR, and FR in Fig. \ref{fig:heat_field_plots}. Furthermore, all employed time integration methods show near identical results as per Tab. \ref{tab:result_overview} as well as Fig. \ref{fig:heat_ti_inference}, which depicts behavior over time, including residuals. Fig.~\ref{fig:heat_example} illustrates inference using explicit Euler, alongside the squared difference between the reference and prediction, as well as the calculated residual on the network's output. As hypothesized, the squared difference and the residual exhibit similar patterns, highlighting the potential of using the residual as an indicator of prediction quality. This relationship becomes even more pronounced when computing the mean absolute error (MAE) and the Pearson correlation coefficient~$\rho$ over all test examples, yielding a MAE of $2.04\eminus3$ and $\rho = 0.997$, thereby indicating an almost perfect correlation across the dataset.

\begin{figure}[tbhp]
    \centering
    \includegraphics[width=\linewidth]{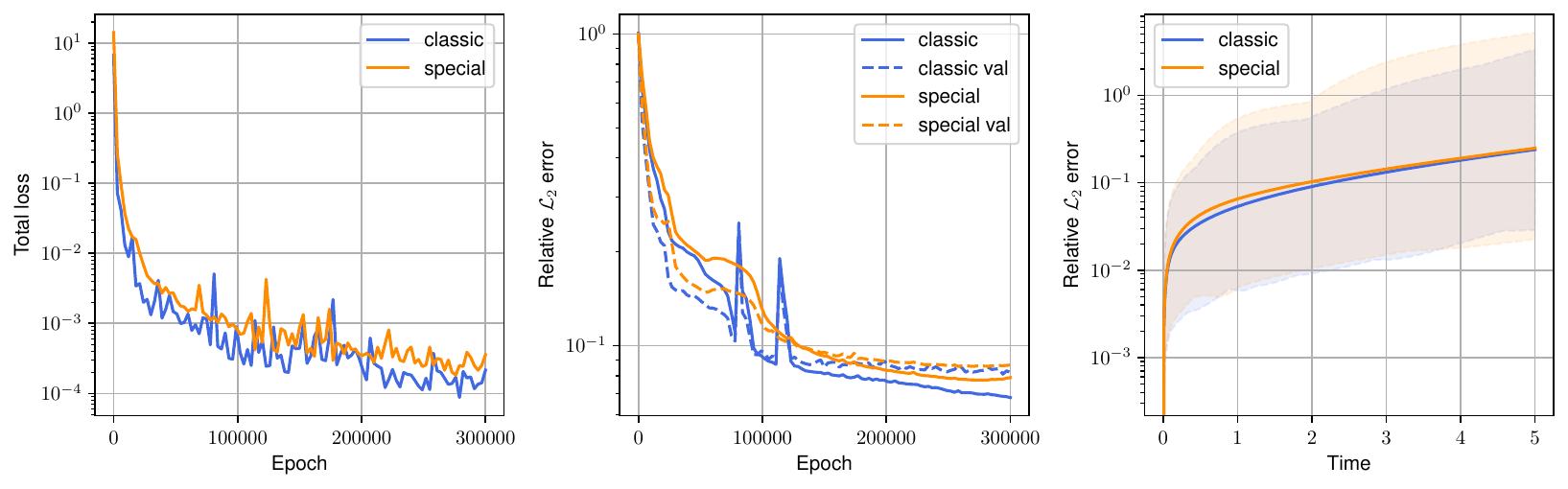}
    \caption{1D Heat equation: Comparison of time integration models in the vanilla form and special form for time-evolution partial differential equations, including total loss and mean relative $\mathcal{L}_2$ error on train and validation set during training. Only the relative $\mathcal{L}_2$ error on the time derivative is provided here. The right plot shows the performance during inference with the fully trained model using an explicit Euler scheme while depicting the minimum, mean, and maximum value of the relative $\mathcal{L}_2$ error per time step.}
    \label{fig:heat_special}
\end{figure}

As mentioned in Section~\ref{sec:training_setup}, a special formulation can be employed for time-evolution PDEs, such as the heat equation, where the time derivative appears in isolation, and all other terms can be moved to the right-hand side. \revresponse{In our experiments, this efficiency-oriented variant yields comparable accuracy} to the standard (vanilla) formulation, with mean relative $\mathcal{L}_2$ errors of $2.39\eminus1$ for the standard and $2.48\eminus1$ for the special formulation (Fig.~\ref{fig:heat_special}).
\subsection{One-dimensional Burgers' Equation}
After examining heat conduction, the Burgers' equation is introduced as the next example, highlighting the transition from diffusion-dominated to hyperbolic dynamics, incorporating nonlinear advection and viscosity effects in fluid flow. The equation reads as: 
\begin{equation}
\label{eq:burgers}
    \frac{\partial u}{\partial t} = 0.01\frac{\partial^2 u}{\partial x^2} - u\frac{\partial u}{\partial x},\quad (t,x)\in [0,\infty)\times[0,1]
\end{equation}
with periodic BCs $u(t,0)=u(t,1)$ and $\tfrac{\partial u}{\partial x}(t,0)=\tfrac{\partial u}{\partial x}(t,1)$. The IC \(u(0,x)=f(x)\) was sampled from a Gaussian process with spectral density $S(k) = \sigma^2\bigl(\tau^2 + (2\pi k)^2\bigr)^{-\gamma}, \quad \sigma=25,\;\tau=5,\;\gamma=4, $ periodic on $x\in[0,1]$. Equivalently, the covariance kernel is $K(\mathbf x,\mathbf x') \;=\; \int_{-\infty}^{\infty} S(k)\,e^{2\pi i k(\mathbf x - \mathbf x')}\,\mathrm{d}k$. A spatial resolution of 101 grid points was used. 

The dataset is generated by solving Eq. \ref{eq:burgers} on a periodic spatial domain $x \in [0, 1]$. The spatial domain is discretized using $s = 4,096$ grid points, while outputs are captured at $n_{\text{steps}} + 1 = 101$ evenly spaced time steps over $t \in [0, 1]$. The ICs $u_0(x)$ are sampled from stationary Gaussian random fields (GRFs) with covariance kernel $C = \sigma^2 (-\Delta + \tau^2 I)^{-\gamma}$. For GRF generation, the Fourier coefficients are drawn from $\mathcal{N}(0, \sigma^2)$ and scaled appropriately to enforce the specified covariance structure, yielding smooth, periodic initial profiles. The Burgers' equation is advanced in time using an operator-splitting spectral method implemented with the \texttt{spin} function from Chebfun~\cite{chebfun}, which combines linear evolution via viscosity $\nu u_{xx}$ with nonlinear convection $-uu_x$. The solution $u(t, x)$ is represented as a Chebyshev interpolant at each time step and stored. The corresponding time derivative $\partial_t u(t, x)$ is independently computed at each output time by substituting the solution into the original Burgers' equation. Consistency checks are performed to ensure the accuracy of $\partial_t u(t, x)$ using both finite differences and the governing PDE. The dataset consists of $N = 2,500$ realizations of $u_0(x)$, where each realization evolves under the Burgers' dynamics and is evaluated on a reduced grid of $n = 101$ spatial points.

\begin{figure}[tbhp]
    \centering
    \includegraphics[width=\linewidth]{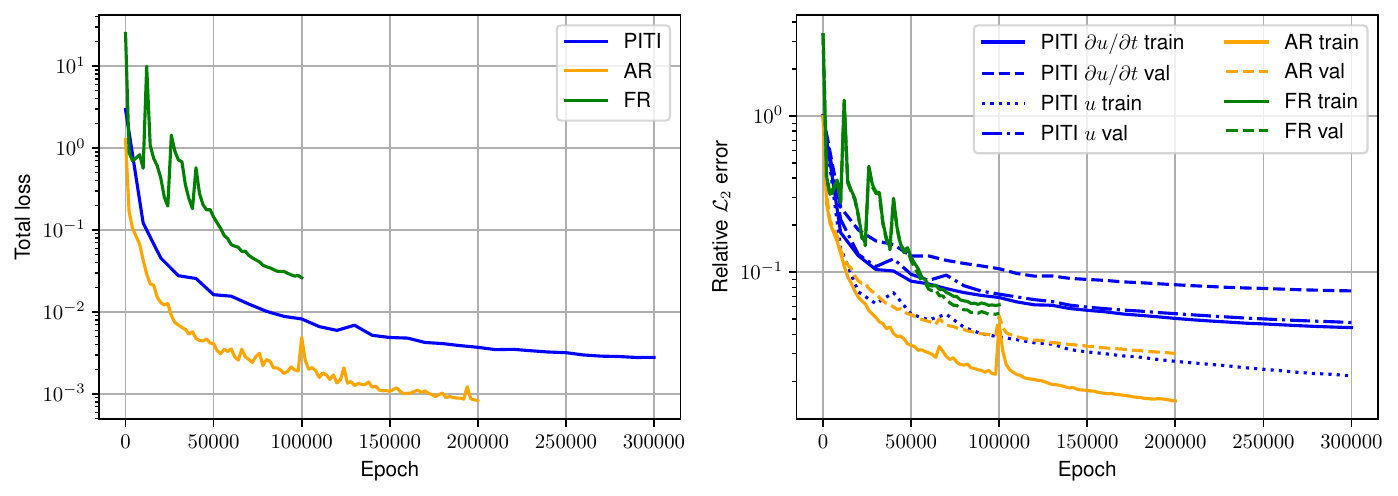}
    \caption{1D Burgers' equation: Total loss during training alongside mean relative $\mathcal{L}_2$ error on the train and validation set for full rollout (FR), autoregressive (AR), and time integration (PITI) model. The relative $\mathcal{L}_2$ error on both network outputs is provided for the PITI model. Hyperparameters were adapted per model as per table \ref{tab:hyperparameters}.}
    \label{fig:burgers_loss}
\end{figure}

In this case, we utilize the hybrid setup by considering 1,200 IC profiles sampled at $t=0$ for purely physics-informed training~(TI at $t=0$, AR with $\rm{dt}=0.01$, FR on $t\in[0, 0.5]$) and additionally 1,200 profiles sampled from 400 trajectories at $t=\lbrace0, 0.25, 0.5\rbrace$ used for both physics-informed and data-driven training. Once again, this data could also be sampled randomly with consistency of $u$ and $\frac{\partial u}{\partial t}$, but sampling from trajectories ameliorates the sampling. The data-driven losses were evaluated for PITI using $\frac{\partial u}{\partial t}$ at $t=0$ by flattening, AR uses $u$ at $t=0, \rm{dt}=0.01$ by using the sampled and next step, and FR uses $u$ at $t = \lbrace0, 0.25, 0.5\rbrace$. Final mean relative $\mathcal{L}_2$ error was $4.41\eminus2$ on the train and $7.56\eminus2$ on the validation set for the time derivative using TI, whereas AR was $1.50\eminus2$ and $3.02\eminus2$, and FR was $6.16\eminus2$ and $5.44\eminus2$ on the field directly~(See Fig. \ref{fig:burgers_loss} for full trajectories). 
\begin{figure}[!htb]
    \centering
    \includegraphics[width=0.49\linewidth]{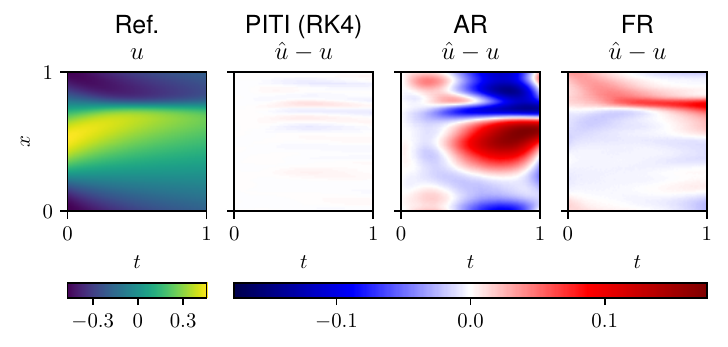}
    \caption{1D Burgers' equation: Reference and difference of predictions to reference using time integration with fourth-order Runge-Kutta, autoregressive and full rollout.}
    \label{fig:burgers_ex_compare}
\end{figure}
\begin{figure}[tbhp]
    \centering
    \includegraphics[width=0.49\linewidth]{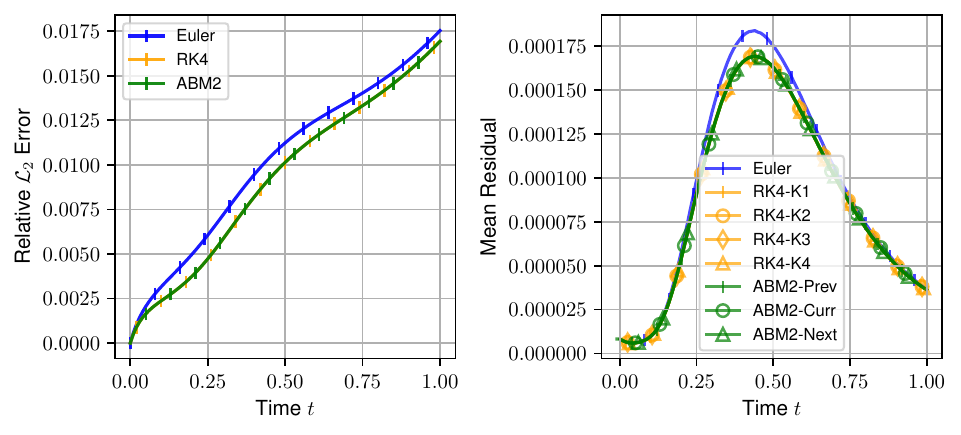}
    \caption{1D Burgers' equation: Comparison of different time integration (PITI) methods using explicit Euler, fourth-order Runge-Kutta (RK4), and second-order Adam-Bashforth-Moultons (ABM2) with mean relative $\mathcal{L}_2$ error on the left and predicted residuals on the right over all test examples. \revresponse{Note that in both subplots several curves overlap almost exactly and therefore appear as a single trace.}}
    \label{fig:burgers_ti_inference}
\end{figure}
\begin{figure}[tbhp]
    \centering
    \includegraphics[width=0.49\linewidth]{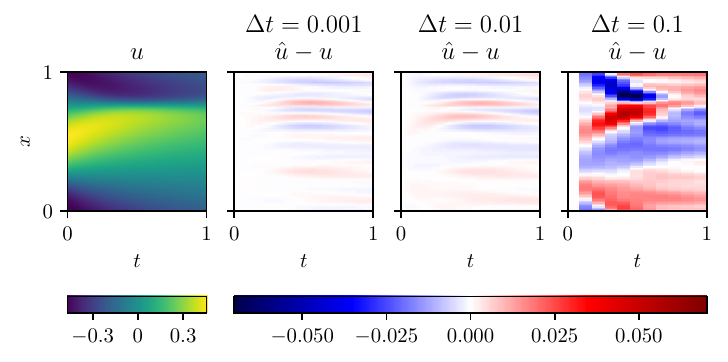}
    \caption{1D Burgers' equation: Reference and difference of predictions to reference using time integration with explicit Euler and different time steps.}
    \label{fig:burgers_dt_compare}
\end{figure}
\begin{figure}[tbhp]
    \centering
    \includegraphics[width=\linewidth]{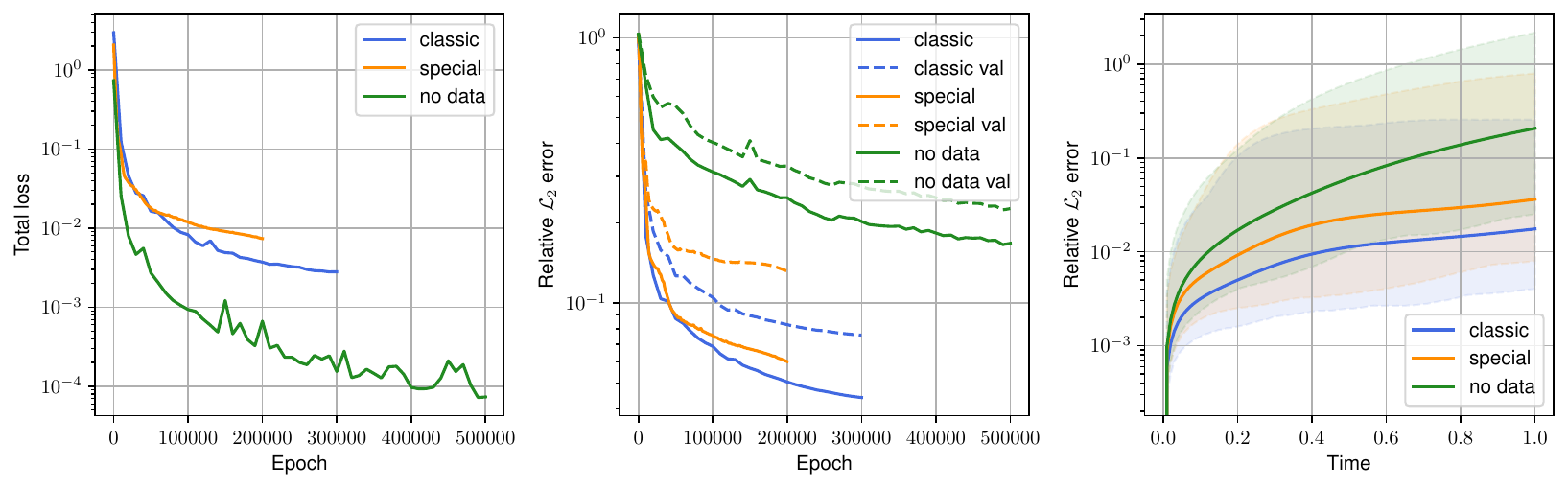}
    \caption{1D Burgers' equation: Comparison of time integration models in the vanilla form and special form for time-evolution partial differential equations alongside a purely physics-informed training including total loss and mean relative $\mathcal{L}_2$ error on train and validation set during training. Only the relative $\mathcal{L}_2$ error on the time derivative is provided here. The right plot shows the performance during inference with the fully trained model using an explicit Euler scheme while depicting minimum, mean, and maximum value of the relative $\mathcal{L}_2$ error per time step.}
    \label{fig:burgers_special}
\end{figure}
As the values are now in the same range, with a slight advantage for AR, it comes as no surprise that PITI clearly yields the best results, as shown in Fig.~\ref{fig:burgers_l2} and Tab.~\ref{tab:result_overview}. Explicit Euler provides slightly worse results compared to RK4 and ABM2 during inference with PITI models (Fig.~\ref{fig:burgers_ti_inference} for full trajectories). Fig.~\ref{fig:burgers_ex_compare} illustrates the difficulty of AR and FR in accurately capturing the shock formation. A further study is presented in Fig. \ref{fig:burgers_dt_compare} showing inference with different timesteps, achieving mean relative $\mathcal{L}_2$ error at the end of testing domain of $1.69\eminus2$ for $\Delta t=0.001$, $1.75\eminus2$ for $\Delta t=0.01$, and $4.86\eminus2$ for $\Delta t=0.1$, where the first satisfies the CFL condition. This becomes evident as the largest time step fails to capture the full dynamics, while the middle time step, despite exceeding the CFL condition, still exhibits stable performance. 

\begin{figure}
    \centering
    \includegraphics[width=0.49\linewidth]{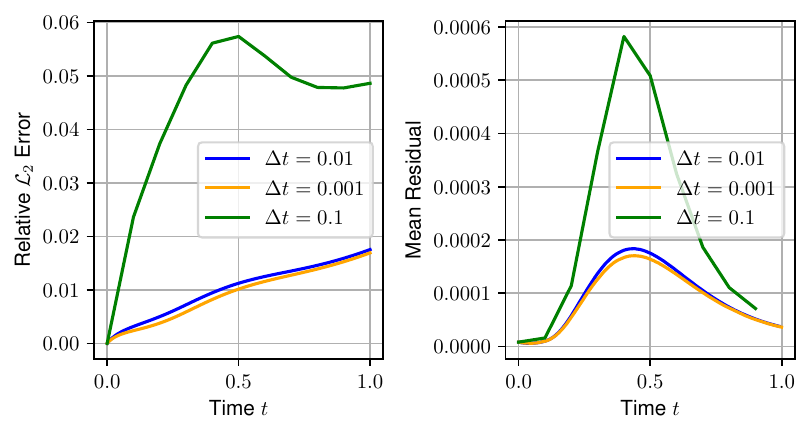}
    \caption{1D Burgers' equation: Comparison of explicit Euler as time integration (TI) method with different time steps ($\Delta t$) during inference. Mean relative $\mathcal{L}_2$ error is shown on the left and predicted residuals on the right over all test examples.}
    \label{fig:burgers_ti_dt}
\end{figure}

The full trajectories as well as the residuals obtained during inference are shown in Fig. \ref{fig:burgers_ti_dt}, with full field plots available in the Fig. \ref{fig:burgers_dt_compare}. The residuals for different time steps~(Fig. \ref{fig:burgers_ti_dt}) as well as different methods~(Fig. \ref{fig:burgers_ti_inference}) indicate a regime around the middle of the testing domain that was learned less accurately compared to states before and after. This is discernible due to an increase in the mean residual over all test examples, followed by a reduction back to smaller residual values. One could exploit this to adaptively resample profiles from this region of increased residuals in order to reduce the overall error. The MAE between the squared difference and residual for explicit Euler with $\Delta t=0.01$ is $2.63\eminus{-5}$ with Pearson's correlation coefficient $\rho=0.9996$, once again suggesting that this correlation can be leveraged to obtain higher quality predictions. The last study for this example includes the already displayed classic form in a hybrid setting against the special form for \revresponse{explicit} time-evolution PDEs~(mean relative $\mathcal{L}_2$ error $ = 3.62\eminus2$) alongside the classic form without additional data~(mean relative $\mathcal{L}_2$ error $2.07\eminus1$). \revresponse{As shown in Fig.~\ref{fig:burgers_special}, the special formulation (cf. Section~\ref{sec:training_setup}) does not provide an accuracy benefit for the more nonlinear or advection-dominated Burgers' equation, and the vanilla formulation remains more robust.} Thus, the chosen setup outperforms the data-free setting by more than an order of magnitude while still halving the error compared to the special case, thus indicating better performance on more complex equations.

\begin{figure}[tbhp]
    \centering
    \includegraphics[width=0.33\linewidth]{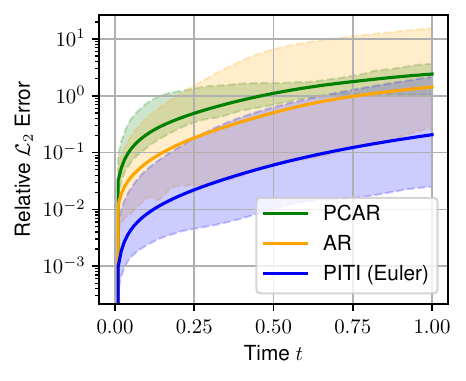}
    \caption{\revresponse{1D Burgers' equation: Time evolution of the relative $\mathcal{L}_2$ error for PCAR, AR, and PITI with explicit Euler in a purely physics-informed setting. The PITI ansatz reduces error growth relative to both auto-regressive rollouts.}}
    \label{fig:Burgers_pcar}
\end{figure}

\revresponse{Since Burgers' equation has been extensively studied in the literature, we provide two additional comparisons to established baselines. First, we consider physics-constrained deep auto-regressive networks (PCAR, \cite{Geneva2020}). We adopt the hyperparameter recommendations of \cite{Geneva2020}, but modify the architecture to match our setting by using a single input slice. We benchmark this adapted PCAR model against a physics-informed DeepONet AR approach and PITI-DeepONet with an explicit Euler time integrator. All three methods are trained under identical conditions with purely physics-informed training using the same data and the same hyperparameters specified above. Test-set results are summarized in Fig.~\ref{fig:Burgers_pcar}. Overall, PCAR performs worse than DeepONet-based AR and PITI. While PCAR achieves a final relative $\mathcal{L}_2$ error of $2.95\eminus2$ on the held-out validation set (computed over all slices across the full time horizon $t\in[0,1]$), its rollout yields a mean relative $\mathcal{L}_2$ error of $2.41$ on the final slice of the test domain (min: $1.03$, max: $3.66$). In contrast, AR attains $1.43$ (min: $2.78\eminus1$, max: $1.69$), and PITI achieves $2.06\eminus1$ (min: $2.52\eminus2$, max: $2.16$).}

\revresponse{Second, we compare against the reported MSE in PINODE~\cite{Sholokhov2023}.PINODE samples initial conditions from a finite harmonic family and considers 128 spatial points for $x\in[-\pi,\pi]$ and $t\in[0,2]$ with $\Delta t=0.1$ in training. In contrast, our setup samples initial conditions from a Gaussian random field, uses $x\in[0,1]$ with 101 spatial points, and considers $t\in[0,1]$ with 101 time steps. While this differs from our setup, PINODE reports MSE values for unrolling beyond the training domain; hence, we include this as a reference. We reference the reported mean MSE in the first post-training period as $7\eminus2$ (physics-only) and $5\eminus3$ (hybrid) as as no official PINODE code is available for a direct re-implementation. While recognizing that the underlying data production and domains differ between the two studies, PITI with explicit Euler achieves $1.1\eminus3$ (physics-informed) and $3.3\eminus5$ (hybrid) in the post-training phase under the same period-based MSE, suggesting significantly lower error in this situation.}

\subsection{Two-dimensional Allen-Cahn Equation}
The final example we consider is the two-dimensional Allen–Cahn equation, governed by:
\begin{equation}
\label{eq:ac}
    \frac{\partial u}{\partial t} = 0.05^2 \left( \frac{\partial^2 u}{\partial x^2} + \frac{\partial^2 u}{\partial y^2} \right) - \left( u^3 - u \right)
\end{equation}
for $(t,x,y)\in [0,\infty)\times[0,1]\times[0,1]$ with periodic BCs $u(t,0,y)=u(t,1,y)$ and $u(t,x,0)=u(t,x,1)$. This PDE is used to investigate phase–field dynamics on a $16\times16$ spatial grid. The IC $u(0,x,y)=f(x,y)$ was sampled from a smooth periodic Gaussian random field (correlation length = 0.1) with linear mapping to $f(x,y)\in [-1,1]$.

The dataset is generated by solving Eq. \ref{eq:ac} on the periodic spatial domain discretized using a uniform $16 \times 16$ grid. The temporal evolution is computed up to a final time $T_{\mathrm{final}} = 5.0$ using an exponential time-differencing fourth-order Runge–Kutta (ETDRK4) scheme. The coarse time step $\Delta t = 0.01$ is subdivided into 200 finer steps, resulting in a fine time resolution $\Delta t_{\mathrm{fine}} = \Delta t / 200$. For spatial discretization, the Laplace operator is represented in spectral space using wavenumbers derived from a Fourier transform grid. Initial conditions are sampled from Gaussian-filtered random fields, smoothed by a low-pass filter, and mapped to the range $[-1, 1]$. During the simulation, the solution $u(t, x, y)$ and its time derivative $\partial_t u(t, x, y)$ are recorded on a coarse temporal grid spanning $n_{\mathrm{coarse}} = T_{\mathrm{final}} / \Delta t + 1$ snapshots. Additionally, the real-space energy functional $E = \int \left( \frac{\varepsilon^2}{2} |\nabla u|^2 + \frac{1}{4} (u^2 - 1)^2 \right) \, \mathrm{d}x \, \mathrm{d}y$ is computed at each snapshot to monitor consistency and stability. The dataset consists of $N_{\mathrm{train}} = 2,000$ training samples and $N_{\mathrm{test}} = 500$ test samples, with unique ICs for each sample. 

\begin{figure}[tbhp]
    \centering
    \includegraphics[width=\linewidth]{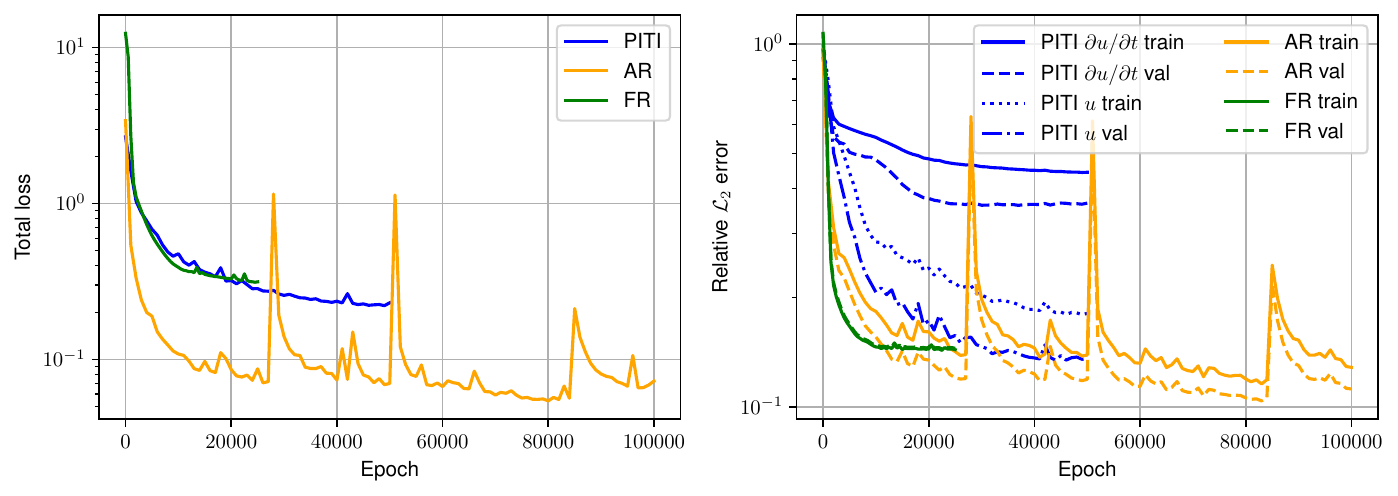}
    \caption{2D Allen-Cahn equation: Total loss during training alongside mean relative $\mathcal{L}_2$ error on the train and validation set for full rollout (FR), autoregressive (AR), and time integration (PITI) model. The relative $\mathcal{L}_2$ error on both network outputs is provided for the PITI model. Hyperparameters were adapted per model as per table \ref{tab:hyperparameters}.}
    \label{fig:ac_loss}
\end{figure}

The same hybrid setup employed in the 1D Burgers' case is utilized here: 400 validation and 500 test samples on $t \in [0,1]$ with $\mathrm{dt} = 0.01$, 1,200 profiles at $t=0$ for physics-informed training, and 1,200 data points sampled from 400 trajectories at $t = \lbrace 0, 0.25, 0.5 \rbrace$. The branch network for this setup was a convolutional neural network (details in Tab.~\ref{tab:hyperparameters}). The final mean relative $\mathcal{L}_2$ error for the time derivative using PITI was $4.43\eminus 1$ on the training set and $3.64\eminus 2$ on the validation set, whereas AR yielded $1.28\eminus 1$ and $1.12\eminus 1$, and FR incurred $1.44\eminus 1$ and $1.45\eminus 1$ on the field directly (see Fig.~\ref{fig:ac_loss} for full trajectories). Once again, although PITI exhibits a higher error on the traced time derivative metric compared to the error on the field itself, it significantly outperforms AR and FR, as shown in Tab.~\ref{tab:result_overview} and Fig.~\ref{fig:ac_l2}, where the upper limit for PITI using RK4 consistently surpasses the mean performance of other methods.

\begin{figure}[htbp]
\begin{subfigure}[b]{\textwidth}
    \centering
    \includegraphics[width=\linewidth]{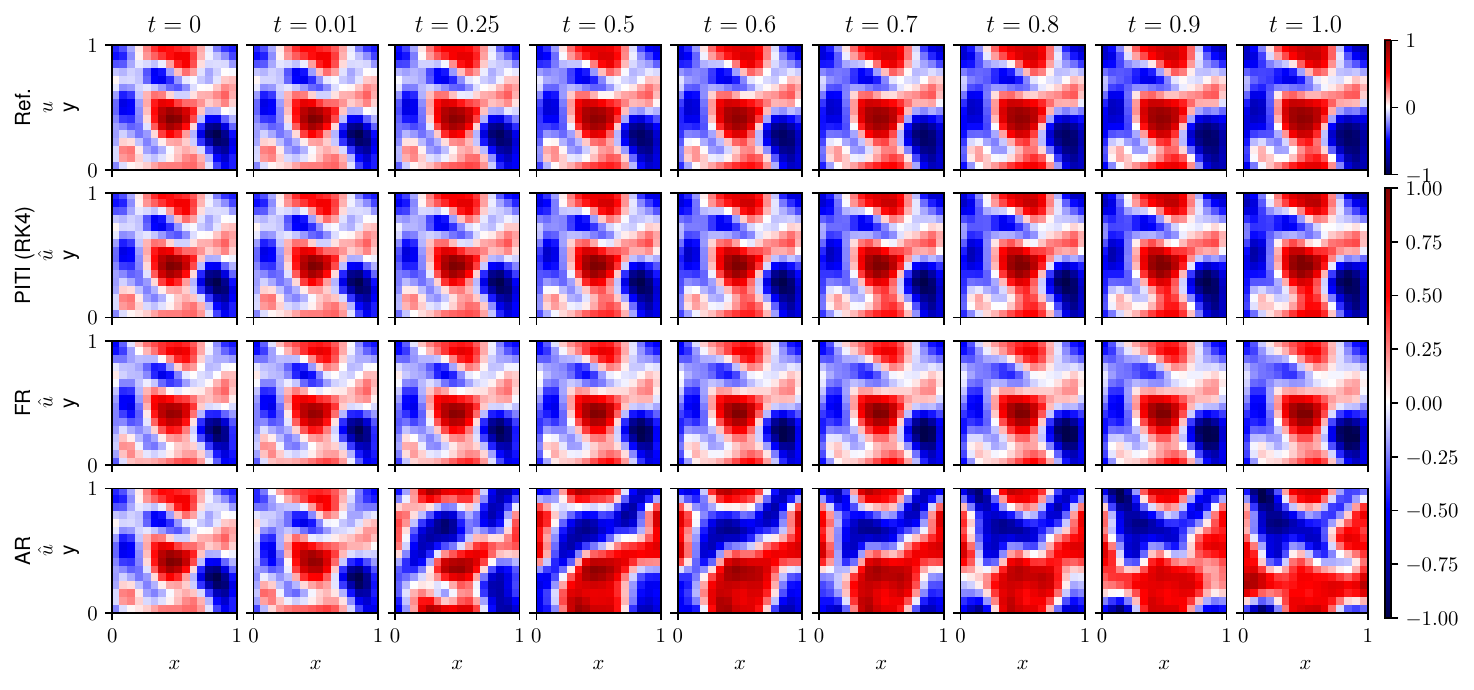}
    \caption{Reference and predictions}
    \label{fig:ac_field_ex}
    \vspace{0.6em}
\end{subfigure}
\begin{subfigure}[b]{\textwidth}
    \centering
    \includegraphics[width=\linewidth]{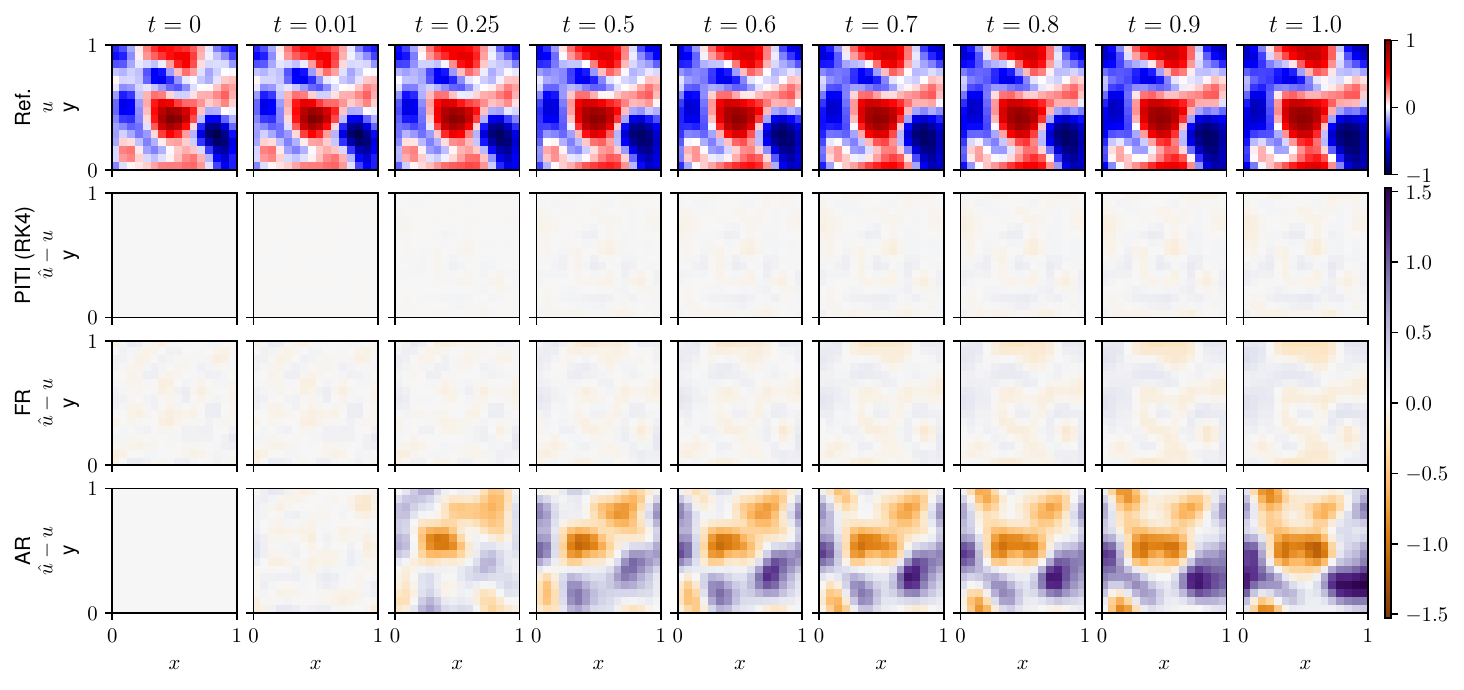}
    \caption{Reference and absolute error in predictions with respect to reference}
    \label{fig:ac_compare}
\end{subfigure}
\caption{2D Allen-Cahn equation: Reference, predictions, and absolute error in predictions with respect to reference using time integration with fourth-order Runge-Kutta~(RK4), autoregressive and full rollout at different time points.}
\label{fig:ac_combined}
\end{figure}

\begin{figure}[tbhp]
    \centering
    \includegraphics[width=0.49\linewidth]{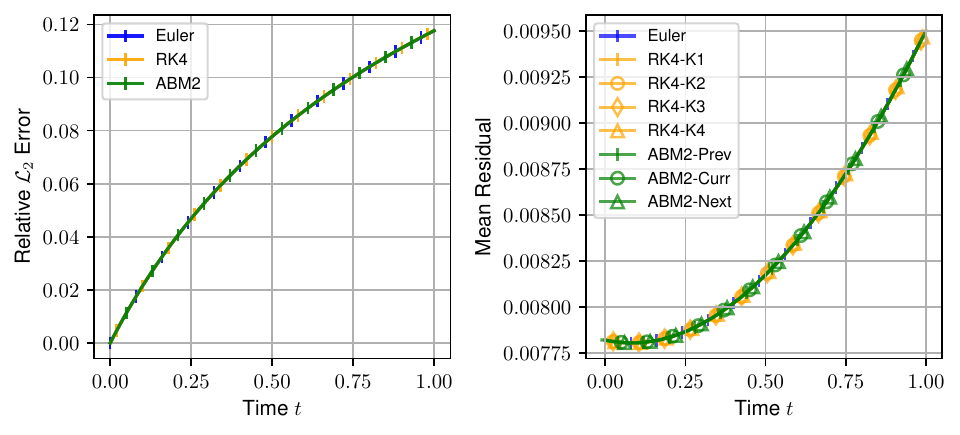}
    \caption{2D Allen-Cahn equation: Comparison of different time integration (PITI) methods using explicit Euler, fourth-order Runge-Kutta (RK4) and second-order Adam-Bashforth-Moultons (ABM2) with mean relative $\mathcal{L}_2$ error on the left and predicted residuals on the right over all test examples. \revresponse{Note that in both subplots several curves overlap almost exactly and therefore appear as a single trace.}}
    \label{fig:ac_ti_inference}
\end{figure}

Furthermore, the chosen time integration scheme during inference has negligible influence on the errors (Tab.~\ref{tab:result_overview} and Fig.~\ref{fig:ac_ti_inference}). An example including error comparison for the methods is shown in Fig.~\ref{fig:ac_compare}, demonstrating the struggle of AR to capture the dynamics even within the training domain, and FR starting to slightly diverge in extrapolation with relatively low dynamics in the remaining field. In contrast, PITI exhibits comparably more gradual divergence, yet its residual continues to grow in the later states, thus hinting at the potential to improve sampling of states to sufficiently represent the underlying dynamics (Fig.~\ref{fig:ac_ti_inference}). Once again, residual and squared difference exhibit near perfect correlation, with an MAE of $1.84\eminus 3$ and Pearson's correlation coefficient $\rho = 0.997$ for explicit Euler with $\Delta t = 0.01$.


\subsection{\revresponse{One-dimensional Kuramoto-Sivashinsky Equation}}
\revresponse{Finally, to demonstrate the applicability of PITI to more chaotic systems, we consider the Kuramoto-Sivashinsky~(KS) equation. The equation reads as:
\begin{equation}
\label{eq:ks}
    \frac{\partial u}{\partial t} = - \frac{\partial^2 u}{\partial x^2} - \frac{\partial^4 u}{\partial x^4}-u\frac{\partial u}{\partial x},\quad (t,x)\in [0,\infty)\times[0,22]
\end{equation}
with periodic BCs $u(t,0)=u(t,22)$ and $\tfrac{\partial u}{\partial x}(t,0)=\tfrac{\partial u}{\partial x}(t,22)$. The initial condition $u(0,x)=f(x)$ was generated by drawing a random periodic profile $u_0(x)$ from i.i.d.\ Gaussian coefficients and evolve the PDE using the Fourier spectral \texttt{spin} solver in \texttt{Chebfun}, with an internal step-size cap $\Delta t_{\max}=10^{-3}$. After a burn-in of $t_{\mathrm{burn}}=23$ time units, which mitigates startup transients, we set $f(x)=u(t_{\mathrm{burn}},x)$ as effective IC thus yielding $f$ representative of the KS long-time statistically stationary regime. From this state, we record $101$ uniformly spaced snapshots over $T=10$ time units and evaluate the solution on a reduced grid of $n=128$ spatial points. 
Time derivatives $\partial_t u(t,x)$ are stored alongside $u(t,x)$ (computed from the KS right-hand side) and cross-checked on the test set using finite differences. In total, the dataset consists of $N_{\mathrm{train}}=2000$ training and $N_{\mathrm{test}}=500$ test trajectories.}

\begin{figure}[tbhp]
    \centering
    \includegraphics[width=0.49\linewidth]{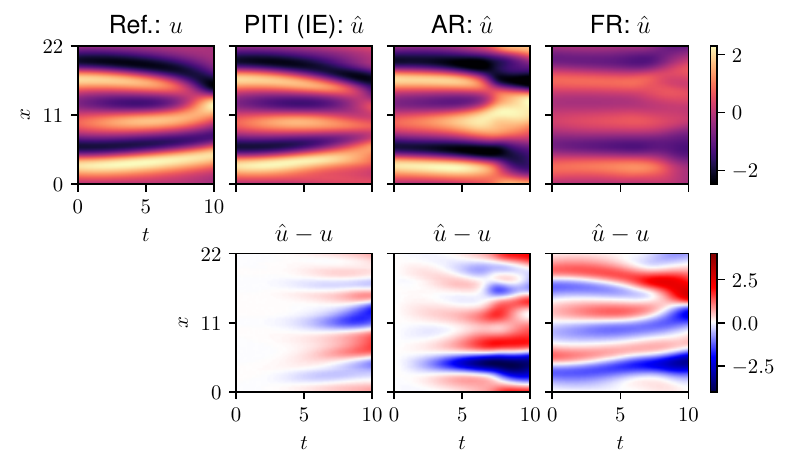}
    \caption{\revresponse{1D Kuramoto-Sivashinsky equation: Reference and difference of predictions to reference using time integration with implicit Euler.}}
    \label{fig:ks_example}
\end{figure}

\revresponse{We implement the hybrid versions for FR, AR, and PITI with 1,200 IC profiles sampled at $t=0$ for purely physics-informed training~(TI at $t=0$, AR with $\rm{dt}=0.01$, FR on $t\in[0, 5.0]$) and additionally 1,200 profiles sampled from 400 trajectories at $t=\lbrace0, 2.5, 5.0\rbrace$ used for both physics-informed and data-driven training. This is the same setup as for Burgers' equation, but adapted to the longer time period for KS. Apart from that, we restrict our baselines to explicit Euler and additionally include backward Euler for stability, while omitting ABM2 and RK4 as inference schemes for PITI. Fig. \ref{fig:ks_l2} depicts a comparison of the relative $\mathcal{L}_2$ error over time, showcasing the inability of FR to capture any dynamics with the given information, while AR deteriorates about a magnitude faster than PITI, which is also discernible in Fig \ref{fig:ks_example}.}

\begin{figure}[tbhp]
    \centering
    \includegraphics[width=0.49\linewidth]{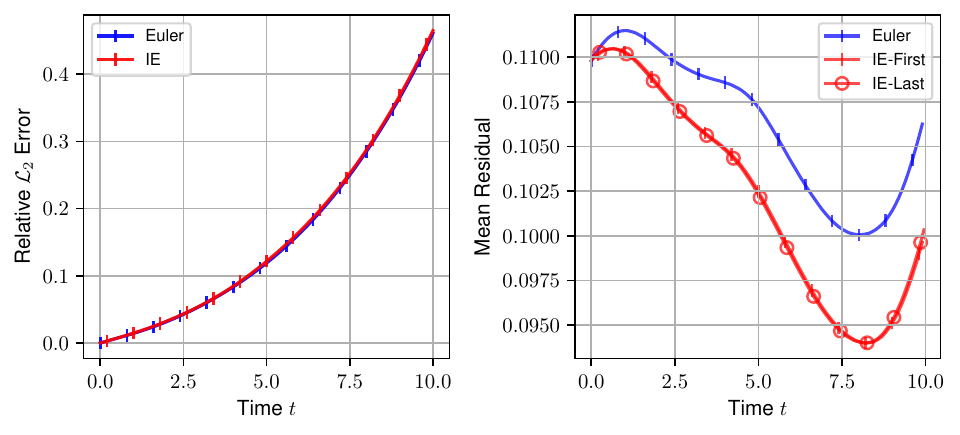}
    \caption{\revresponse{1D Kuramoto-Sivashinsky equation: Reference and difference of predictions to reference using time integration with implicit Euler. Only the first and last residual in each iteration for IE is shown.}}
    \label{fig:ks_l2_residual}
\end{figure}

\revresponse{While the resulting relative $\mathcal{L}_2$ errors were comparable, the reconstruction errors during inference were lower for implicit Euler as given in Fig. \ref{fig:ks_l2_residual}. This is particularly interesting as the underlying operator is identical, and the numerical update seems to keep the state within the learned domain for the A-stable implicit Euler. Once again, linear correlation between reconstruction error and error in the time derivative is almost perfect with $\rho=0.974$. This slightly worse value compared to the other three methods seems to stem from the more pronounced error on the field value with a relative $\mathcal{L}_2$ error of $2.5\eminus1$ on the held-out validation data compared to the error for the time derivative with $6.8\eminus2$ at the end of training.}

\revresponse{These results for KS showcase the strength of PITI to work for chaotic equations with limited data, where FR fails to capture any underlying dynamics even on the training data within the training domain with a relative $\mathcal{L_2}$ errors of 0.69. AR was able to capture this dynamic with relative $\mathcal{L}_2$ errors of $1.63\eminus2$ on the train and $1.32\eminus2$ on the validation data at the end of training, but during inference, the predicted trajectory rapidly diverges from the reference solution. Even though the employed time-integration schemes are not standard choices for the stiff, chaotic Kuramoto–Sivashinsky equation, they still deliver markedly better performance than the reference methods using operator learning in a hybrid physics-informed and limited data-driven setup.}

\subsection{Ablation Studies}
\revresponse{To further characterize the robustness of the results, we investigate how random initialization, training-set size, and the inference time step size influence the reported performance.}
\begin{table}[tbph]
    \centering
    \caption{Seed sensitivity analysis for the final model configurations. Each method (PITI, AR, FR) was trained with five different random seeds. For each run, the relative $\mathcal{L}_2$ error was computed over all test samples and summarized by its mean, minimum, and maximum at two time points corresponding to the end of the training domain and the end of the testing domain for each problem (Heat 1D: $t=0.5$ and $t=5.0$; Burgers 1D: $t=0.5$ and $t=1.0$; Allen-Cahn 2D: $t=0.5$ and $t=1.0$\revresponse{; Kuramoto-Sivashinsky~(KS) 1D: $t=5.0$ and $t=10.0$}). Table entries report the overall mean and between-seed standard deviation across the five runs, formatted as $\mu \pm \sigma$.}
    \vskip1.25mm
    \footnotesize{
    \resizebox{\columnwidth}{!}{
    \begin{tabular}{@{}ll|cc|cc|cc|cc@{}}
\toprule
&      & \multicolumn{2}{c|}{Heat 1D}          & \multicolumn{2}{c|}{Burgers 1D}         & \multicolumn{2}{c}{Allen-Cahn 2D}& \multicolumn{2}{c}{\revresponse{KS 1D}} \\ \midrule
&      & $t=0.5 $            & $t=5.0$             & $t=0.5$               & $t=1.0$             & $t=0.5 $            & $t=1.0$   & \revresponse{$t=5.0$}            & \revresponse{$t=10.0$}         \\ \midrule
\multirow{3}{*}{\rotatebox{90}{\strut Mean}} & PITI & $3.5\eminus2\pm3.9\eminus3$ & $2.2\eminus1\pm3.4\eminus2$ & $1.1\eminus2\pm6.2\eminus4  $ & $1.8\eminus2\pm1.8\eminus3$ & $7.9\eminus2\pm1.4\eminus3$ & $1.2\eminus1\pm2.6\eminus3$ & \revresponse{$1.1\eminus1\pm 9.5\eminus3$} & \revresponse{$4.2\eminus1\pm 3.4\eminus2$} \\
                     & AR   & $2.2\eminus1\pm3.3\eminus2$ & $1.5\mathrm{e}0\pm5.4\eminus1$ & $2.4\eminus1\pm8.7\eminus2  $ & $6.0\eminus1\pm2.1\eminus1$ & $9.9\eminus1\pm7.2\eminus2$ & $1.1\mathrm{e}0\pm7.0\eminus2$ & \revresponse{$8.8\eminus1\pm 2.8\eminus1$} & \revresponse{$1.2\mathrm{e}0\pm 4.5\eminus2$} \\
                     & FR  & $5.7\eminus3\pm3.2\eminus4$ & $1.3\mathrm{e}0\pm1.4\eminus1 $ & $3.5\eminus2\pm2.0\eminus3  $ & $1.2\eminus1\pm2.1\eminus2$ & $1.3\eminus1\pm2.3\eminus2$ & $2.1\eminus1\pm1.8\eminus2$ & \revresponse{$6.9\eminus1\pm 6.1\eminus2$} & \revresponse{$1.0\mathrm{e}0\pm 7.5\eminus2$}\\ \midrule
\multirow{3}{*}{\rotatebox{90}{\strut Min}}   & PITI & $3.8\eminus3\pm4.2\eminus4$ & $2.8\eminus2\pm1.7\eminus3$ & $2.5\eminus3\pm3.9\eminus4  $ & $4.4\eminus3\pm5.9\eminus4$ & $5.3\eminus2\pm1.5\eminus3$ & $7.7\eminus2\pm3.6\eminus3$ & \revresponse{$2.9\eminus2\pm 3.8\eminus3$} & \revresponse{$7.5\eminus2\pm 4.7\eminus3$}\\
                     & AR & $3.2\eminus2\pm8.3\eminus3$ & $1.3\eminus1\pm5.9\eminus2$ & $4.1\eminus2\pm7.0\eminus3$ & $1.0\eminus1\pm3.3\eminus2$ & $1.8\eminus1\pm3.6\eminus2$ & $1.9\eminus1\pm4.3\eminus2$ & \revresponse{$2.2\eminus1\pm 1.2\eminus1$} & \revresponse{$3.1\eminus1\pm 6.7\eminus2$}\\
                     &FR  & $1.4\eminus3\pm3.4\eminus4$ & $6.3\eminus1\pm1.3\eminus1$ & $7.0\eminus3\pm7.2\eminus4  $ & $3.9\eminus2\pm1.0\eminus2$ & $4.0\eminus2\pm7.1\eminus3$ & $7.1\eminus2\pm1.3\eminus2$ & \revresponse{$1.2\eminus1\pm 5.2\eminus2$} & \revresponse{$4.9\eminus1\pm 1.5\eminus1$}\\ \midrule
\multirow{3}{*}{\rotatebox{90}{\strut Max}}  & PITI & $1.7\eminus1\pm3.4\eminus2$ & $2.5\mathrm{e}0\pm6.6\eminus1$ & $1.9\eminus1\pm3.7\eminus2  $ & $2.5\eminus1\pm7.1\eminus2$ & $1.1\eminus1\pm6.0\eminus3$ & $1.8\eminus1\pm1.2\eminus2$ & \revresponse{$4.0\eminus1\pm 6.3\eminus2$} & \revresponse{$1.3\mathrm{e}0\pm 1.1\eminus1$}\\
                     & AR  & $9.7\eminus1\pm2.1\eminus1$ & $3.2\mathrm{e}1\pm1.7\mathrm{e}1  $ & $1.8\mathrm{e}0\pm1.0\mathrm{e}0    $ & $5.2\mathrm{e}0\pm3.5\mathrm{e}0  $ & $3.3\mathrm{e}0\pm4.1\eminus1 $ & $2.9\mathrm{e}0\pm3.8\eminus1$ & \revresponse{$2.1\mathrm{e}0\pm 2.5\eminus1$} & \revresponse{$2.4\mathrm{e}0\pm 3.0\eminus1$}\\
                     & FR   & $2.9\eminus2\pm5.9\eminus3$ & $2.1\mathrm{e}1\pm2.5\mathrm{e}0  $ & $2.5\eminus1\pm3.7\eminus2  $ & $5.3\eminus1\pm9.3\eminus2$ & $3.4\eminus1\pm6.8\eminus2$ & $4.7\eminus1\pm6.6\eminus2$ & \revresponse{$1.7\mathrm{e}0\pm 6.1\eminus1$} & \revresponse{$2.2\mathrm{e}0\pm 2.6\eminus1$}\\ \bottomrule
\end{tabular}
    }}
    \label{tab:seedstats}
\end{table}
\subsubsection{\revresponse{Random Seed}}
In addition to the investigations described above and the network architectures explored during hyperparameter optimization, the final configurations were trained using five different random seeds. For evaluation, the relative $\mathcal{L}_2$ error was computed for each test sample, and its mean, minimum, and maximum were recorded at two representative time points: (i) the end of the training time horizon and (ii) the end of the test time horizon. Across the five seeded trainings, we report the grand mean and the between-seed standard deviation~(formatted as $\mu \pm \sigma$) of these summary statistics in Table~\ref{tab:seedstats}.

Overall, the between-seed standard deviations are typically at least one order of magnitude smaller than the corresponding error levels. In the remaining cases, the variability is either noticeably smaller while remaining within the same order of magnitude, or the absolute error is already sufficiently large that seed-induced variation is negligible in comparison. This indicates that the reported performance is robust with respect to the choice of random seed and that seed effects do not materially affect the conclusions drawn in the previous sections.

\subsubsection{\revresponse{Available Data and Initial Conditions}}
\begin{figure}[tbhp]
    \centering
    \includegraphics[width=\linewidth]{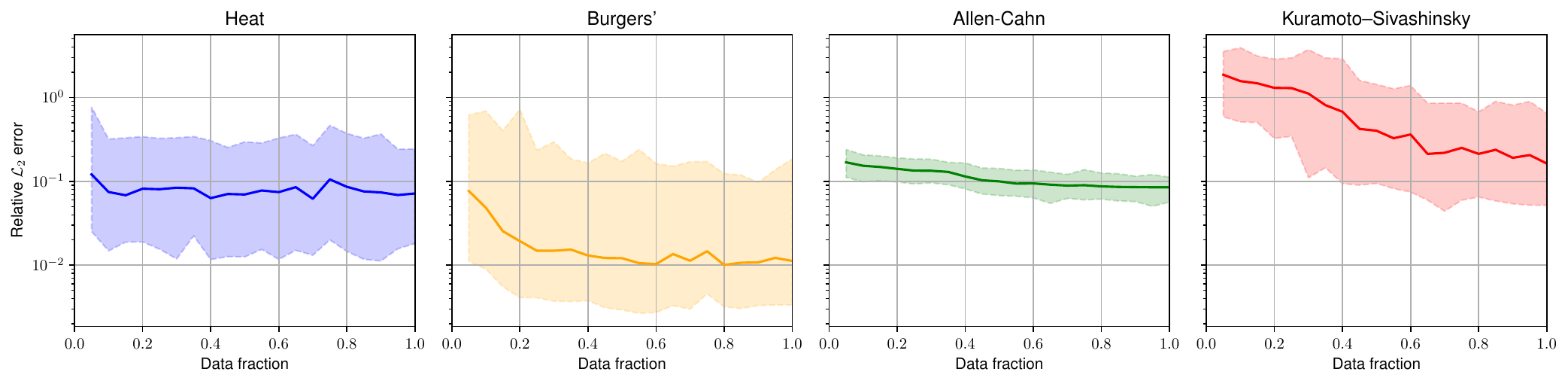}
    \caption{\revresponse{Data ablation study for Heat, Burgers', Allen--Cahn Equation, and Kuramoto-Sivashinsky equation. Models are retrained with fixed hyperparameters while varying the fraction of available training data from $0.05$ to $1.0$ (step $0.05$). Solid lines show the mean relative $\mathcal{L}_2$ error over a fixed test set, and shaded regions indicate min/max across test examples. Inference uses explicit Euler time integration.}}
    \label{fig:data_ablation}
\end{figure}
\revresponse{We retrain the models using the same reported hyperparameters and training budget as in the main experiments while varying the available training data fraction from $0.05$ to $1.0$ in increments of $0.05$ to quantify how performance scales with the number of training samples. All runs are evaluated on the same fixed test set and use explicit Euler time integration during inference. Figure~\ref{fig:data_ablation} reports the mean relative $\mathcal{L}_2$ error across test examples together with the minimum and maximum error observed over the test set.}

\revresponse{The results indicate that the sensitivity to additional training data is strongly benchmark-dependent. For the 1D heat equation, the mean error remains largely unchanged over the full range of data fractions, which is consistent with the diffusion-dominated character of this example and implies that the benchmark is saturated already at small fractions. While the mean performance changes only marginally, the min/max envelope becomes slightly tighter, indicating improved worst-case behavior with additional data but also decreased best-case performance. In contrast, for the 1D Burgers' equation, the mean error decreases more noticeably with increasing data and essentially reaches the final performance already at approximately $60\%$ of the training data, after which further gains are small. Lastly, the error for the higher-dimensional Allen--Cahn example shows signs of convergence toward bigger fractions and drops more gradually over the whole range, reflecting the increased complexity of the dynamics. This response is more pronounced in KS equation, which even indicates that further improvement seems achievable with more data than used within this work.}

\revresponse{For hybrid training configurations, data-driven components are scaled consistently with the available data fraction. Overall, the ablation confirms that the proposed method remains robust even when the training data is significantly reduced.}

\subsubsection{\revresponse{Inference Time Step}}
\begin{figure}[tbhp]
    \centering
    \includegraphics[width=\linewidth]{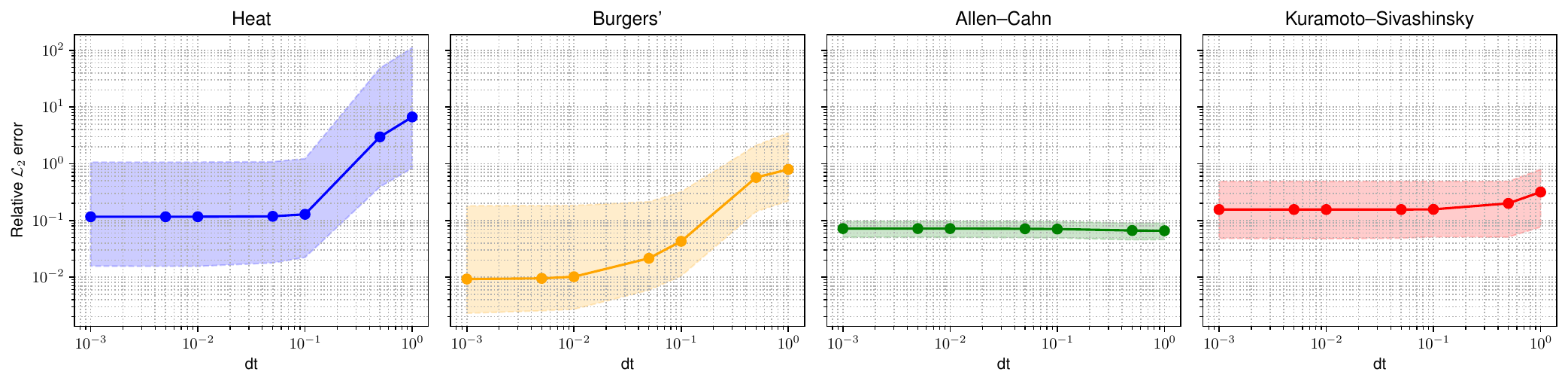}
    \caption{\revresponse{Inference time step analysis using explicit Euler. The models are used in inference with varying step size $\Delta t$. Solid lines show the mean relative $\mathcal{L}_2$ error across test examples, and shaded regions indicate min/max across the test set.}}
    \label{fig:timestep_ablation}
\end{figure}
\revresponse{Since the proposed approach advances the solution by time integrating a learned temporal tangent operator, we assess robustness with respect to the inference step size $\Delta t$. We note that the manuscript already includes a time-step study for inference in Fig.~\ref{fig:burgers_dt_compare}, where we used a smaller and larger $\Delta t$ for the Burgers' equation, while all other inference results in the manuscript were obtained with $\Delta t=0.01$. Building on this, we sweep the inference step size $\Delta t$ over several orders of magnitude using explicit Euler, without retraining the models, and evaluate all runs on the same fixed test set. We report the minimum, mean, and maximum relative $\mathcal{L}_2$ error across test examples (cf. Fig.~\ref{fig:timestep_ablation}).}

\revresponse{Across all four PDEs, we observe a plateau where decreasing $\Delta t$ does not lead to increased accuracy. This suggests that once $\Delta t$ is sufficiently small, the learned tangent operator is the primary source of error rather than temporal discretization. For coarser steps, the error increases due to loss of temporal resolution, with a sharper deterioration for the heat equation and a more gradual deterioration for Burgers’ equation, consistent with the need to resolve stronger nonlinear/advection-driven dynamics. For Allen-Cahn, the mean error slightly improves for larger $\Delta t$, which is in line with reduced accumulation of small systematic model biases when fewer Euler steps are taken, especially because smaller $\Delta t$ implies many more Euler steps, so small systematic errors in the learned tangent can accumulate over time. Nonetheless, for the Allen–Cahn example, the error is largely invariant with respect to the chosen time step $\Delta t$ over the tested range. Lastly, this behavior is also present for the KS equation, as the method would benefit strongly from a better learned derivative than using a smaller timestep.}

\revresponse{We emphasize that this experiment is intended as a step-size robustness study for the learned dynamics, rather than a classical solver-order convergence test, because the learned operator introduces an error floor and repeated operator evaluations can accumulate model bias when $\Delta t$ becomes very small. Furthermore, the time step $\Delta t=0.01$ in the main part is a good choice for all PDEs, with the potential to use higher time steps for the heat and Allen-Cahn equations. Finally, the absolute error levels differ across PDEs, reflecting differences in the difficulty of learning the temporal tangent operator for the respective state distributions, rather than differences in the time-step size being the limiting factor.}

\section{Discussion}
PITI-DeepONet offers rapid, high-accuracy inference with limited to no training data, as demonstrated by the numerical examples \revresponse{even for equations that are considered chaotic}. Extrapolation using this method proves to be more stable and outperforms AR and FR methods, even extending far beyond the domain that was sampled for training: for the heat equation example up to 10 times the training domain. While the learned temporal tangent space operator incurs a higher inference cost compared to FR, it surpasses these methods in accuracy. We further explored different architectures for learning the temporal tangent space of time-evolution PDEs, but the results are similar or worse, while the presented architecture is the most general, applying to PDEs with mixed and/or higher-order derivatives in time and space. For the latter, one might incorporate an additional consistency loss, although the optimization process may become more challenging. While not the focus of this study, the accuracy of our model depends on the reliability of AD, and it may be affected if AD produces unstable results.

The proposed method includes a built-in mechanism for tracking residuals during field reconstruction, providing an indicative measure of prediction quality at inference time. Across all examples, the residuals exhibited a nearly perfect linear correlation with the squared difference between predictions and the reference solutions obtained using explicit Euler, highlighting the strong error-monitoring capabilities inherent in the proposed PITI-DeepONet framework. This residual tracking approach enables additional possibilities, such as coupling it with more accurate schemes when a residual threshold is exceeded or facilitating active learning by identifying whether a state lies within the training domain. It can also serve as a time-step control criterion, allowing the rejection of a step if an unlearned or unexpected state arises, ensuring consistency in the inferred behavior.

A crucial aspect of effectively employing PITI-DeepONet is ensuring that all potential states encountered during inference are adequately sampled. These states can be obtained through experiments, traditional methods such as numerical or analytical solutions, or alternative surrogate models, such as FR simulations. Incorporating expert domain knowledge can significantly enhance the efficiency and quality of the sampling process. An additional, yet unexplored, avenue is the use of multi-fidelity learning~\cite{howard2023multifidelity}, where data of varying resolutions or qualities can be seamlessly included within this framework. Moreover, it has been shown that a hybrid approach, which directly incorporates observed data, can greatly improve the learning process. However, this comes with the potential requirement of time derivative data for training, which may or may not be readily available, depending on the method used to acquire the data.

It is important to emphasize that no specific assumptions were made regarding the neural network architecture, allowing for the use of flexible, problem-specific designs. This can include architectures tailored to the number of variables, such as multi-input~\cite{jin2022mionet} and multi-output~\cite{Lu2022}, as well as multi-network setups, i.e., a single network per field variable in coupled problems~\cite{Cai2021}. In this work, DeepONets with traditional feed-forward neural networks for the branch and trunk, as well as convolutional neural networks for the branch, were utilized; however, other (physics-informed) NO architectures are equally applicable. Furthermore, this framework can be readily scaled to higher-dimensional PDEs by adopting separable architectures~\cite{Mandl2025}.

\section{Conclusion}
In summary, a practical architecture, PITI-DeepONet, is presented that achieves long-term accuracy with reasonable inference time, enabling near real-time predictions while incurring only a marginal increase in computational cost compared to traditional operator learning methods. The proposed approach leverages physics-informed temporal tangent space learning with restricted sampling, exhibiting strong, reliable extrapolation capabilities. Although the current study focused on \revresponse{four} academic benchmark problems, this method has the potential to transform surrogate operator learning by bridging the gap between data-driven/physics-informed learning and classical numerical solvers. As a natural next step, the approach will be extended to tackle more challenging and potentially chaotic/stiff PDE systems.

\section*{Acknowledgments}
LM and TR were supported by Deutsche Forschungsgemeinschaft~(DFG, German Research Foundation) by grant number 465194077~(Priority Programme SPP 2311,~Project SimLivA) and under Germany's Excellence Strategy – EXC 2075 – 390740016. TR thanks the Deutsche Forschungsgemeinschaft (DFG, German Research Foundation) for support via the project ``Hybrid MOR" by grant number 504766766. TR is further supported by the Federal Ministry of Education and Research (BMBF, Germany) within ATLAS by grant number 031L0304A. LM is supported by the Add-on Fellowship of the Joachim Herz Foundation. The research efforts of DN and SG were partly supported by the National Science Foundation (NSF) under Grant No. 2438193 and 2436738. Any opinions, findings, conclusions, or recommendations expressed in this material are those of the author(s) and do not necessarily reflect the views of the funding organizations.

\section*{Author contributions}
LM, DN, TR, and SG conceptualized this work. LM generated the data, implemented the training and inference schemes in software, and prepared results for analysis. LM, DN, and SG performed the formal analysis of the results. TR and SG supervised the project. LM, DN, and SG drafted the original manuscript, and all authors reviewed and edited the final version.

\bibliographystyle{unsrtnat}  
\bibliography{aaai2026}
\end{document}